\documentclass[10pt,twocolumn,letterpaper]{article}

\usepackage[pagenumbers]{cvpr}
\makeatletter
\patchcmd{\@maketitle}{\vskip .5em}{\vskip\dimexpr .5em-.225in\relax}{}{\errmessage{cvpr.sty title block changed: author space not patched}}
\patchcmd{\@maketitle}{\vspace*{24pt}}{\vspace*{12pt}}{}{\errmessage{cvpr.sty title block changed: title space not patched}}
\makeatother

\usepackage{cuted} %

\usepackage{currfile} %

\usepackage{caption} %
\usepackage{multirow} %
\usepackage{colortbl} %

\makeatletter
\renewcommand\paragraph{\@startsection{paragraph}{4}{\z@}%
  {1.5ex \@plus .5ex \@minus .2ex}{-1em}{\normalfont\normalsize\bfseries}}
\makeatother

\usepackage{pifont}
\newcommand{\cmark}{\ding{51}}
\newcommand{\xmark}{\ding{55}}
\newcommand{\pmark}{$\circ$}

\usepackage{tabularx}
\usepackage{array}
\newcolumntype{L}{>{\raggedright\arraybackslash}X}
\newcolumntype{P}[1]{>{\raggedright\arraybackslash}p{#1}}
\newcommand{\code}[1]{\texttt{\small #1}}

\definecolor{cvprblue}{rgb}{0.21,0.49,0.74}
\usepackage[pagebackref,breaklinks,colorlinks,allcolors=cvprblue]{hyperref}
\hypersetup{pdftitle={UEC-Bench: Benchmarking Unspecified Entity Consistency in Multi-Shot Video Generation},
  pdfauthor={Jinyuan Hu, Yifan Shen, Yichang Jian, Peiyuan Zhu, Yuxin Wang, Guangyi Chen, Kun Zhang}}

\title{UEC-Bench: Benchmarking Unspecified Entity Consistency\\in Multi-Shot Video Generation}

\author{%
Jinyuan Hu$^{1}$\quad
Yifan Shen$^{1}$\quad
Yichang Jian$^{2}$\quad
Peiyuan Zhu$^{1}$\\
Yuxin Wang$^{3}$\quad
Guangyi Chen$^{1,2}$\quad
Kun Zhang$^{2,1}$\\
{\normalsize $^{1}$Mohamed bin Zayed University of Artificial Intelligence\quad
$^{2}$Carnegie Mellon University\quad
$^{3}$Wuhan University}%
}

\begin{document}
\maketitle
\begin{strip}
\centering
\includegraphics[width=\textwidth]{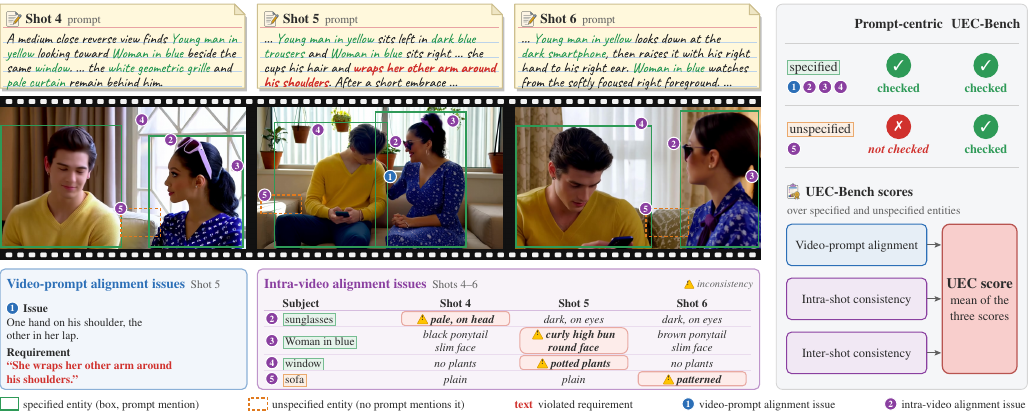}
\captionsetup{hypcap=false}
\captionof{figure}{\textbf{UEC-Bench evaluates entities that no prompt mentions.}
In this episode, Shot 5 violates a prompt requirement, and several entities change across shots.
One of them is the unspecified sofa, which turns from plain to patterned in Shot 6.
Unlike prompt-centric evaluation, which checks only specified entities and misses this change, UEC-Bench scores prompt alignment, intra-shot consistency, and inter-shot consistency over specified and unspecified entities alike and averages the three into the UEC score.}
\captionsetup{hypcap=true}
\label{fig:\currfilebase}%
\end{strip}

\begin{abstract}
Generating coherent multi-shot videos requires more than following textual instructions: visual content must remain consistent across the whole sequence, even when it is not explicitly specified in the prompts.
Yet, existing evaluations largely focus on prompt-specified content, and the consistency of unspecified entities that emerge during generation still lacks automatic, entity-level evaluation.
In this paper, we introduce UEC-Bench (\textbf{U}nspecified \textbf{E}ntity \textbf{C}onsistency Benchmark), a benchmark for systematically evaluating entity consistency in multi-shot video generation, with a particular focus on entities beyond explicit prompt specifications.
UEC-Bench evaluates both prompt-specified and unspecified entities, capturing whether generated videos faithfully follow textual constraints while preserving coherent entity identities, attributes, and relationships both within and across shots.
To enable comprehensive evaluation, we develop an agent-based pipeline that first identifies the entities in the generated videos, including those unspecified in the prompts.
It then detects two kinds of issues, video-prompt alignment issues and intra-video alignment issues, and scores them in three dimensions: prompt alignment for the former, and intra-shot and inter-shot consistency for the latter.
Agent-based assessments are further cross-checked by a dedicated verifier, and the three dimensions are aggregated into a unified score.
Using UEC-Bench, we systematically evaluate a diverse set of multi-shot video generation models, revealing unspecified entity consistency challenges that are overlooked by prompt-centric evaluation.
\end{abstract}

\section{Introduction}
\label{sec:intro}

Recent advances in video generation have enabled models to synthesize increasingly realistic and controllable videos from textual descriptions~\cite{vdm,makeavideo,stablevideodiffusion,yang2025cogvideox,kong2024hunyuanvideo,wan2025wan}.
Building on this progress, multi-shot video generation has emerged as an important direction toward longer and more structured visual narratives, where a sequence of shots jointly depicts a coherent scene or story~\cite{guo2025long,wu2026cinetrans,meng2026holocine,wang2026multishotmaster,zhang2025storymem}.
Compared with single-shot generation, however, multi-shot generation introduces an additional challenge: individual shots should not only faithfully realize their corresponding textual instructions, but also remain visually coherent with one another throughout the sequence.

A fundamental difficulty is that text prompts specify only part of the visual content that eventually appears in a generated video.
A prompt may describe a character, an action, and a scene, while leaving many other visual elements unspecified, such as clothing details, accessories, background objects, or spatial relations.
The generation model must instantiate these unspecified elements to produce complete visual scenes.
Once instantiated, however, they become part of the generated visual context and are often expected to remain coherent within and across subsequent shots.
For example, a background object that emerges in one shot should not arbitrarily change its appearance in the next, even if the prompts never explicitly describe it (Figure~\ref{fig:teaser}).

This distinction exposes an important limitation of existing evaluation paradigms.
Conventional video generation benchmarks primarily assess global video-level properties, including visual quality, temporal dynamics, and text-video alignment~\cite{huang2024vbench,zheng2025vbench,liu2024evalcrafter}.
More recent benchmarks have moved toward fine-grained evaluation at the entity level, assessing whether prompt-specified entities are faithfully generated~\cite{he2026entitybench}.
However, their evaluation is still restricted to a predefined set of entities derived from the prompts or annotations.
As a result, entities that emerge during generation but are not explicitly specified are rarely evaluated automatically at the entity level, despite constituting a substantial part of the generated visual content.

Closing this gap has long been difficult, since it requires identifying real problems about unspecified entities under changing viewpoints and occlusion.
Fortunately, recent progress in vision-language models (VLMs), such as  GPT-6 and Gemini~3.8, has made this difficulty tractable with their strong reasoning and planning.
A straightforward approach is to ask these models to rate each generated video directly.
However, direct rating has many limitations, such as overlooked requirements in long multi-shot prompts, unverified hallucinations, unclear scoring criteria, and incomplete and coarse localization of the underlying problems.
To address these limitations, we introduce UEC-Bench (\textbf{U}nspecified \textbf{E}ntity \textbf{C}onsistency Benchmark), a benchmark designed to evaluate entity consistency beyond explicit prompt specifications.
Rather than restricting evaluation to entities predefined by text, UEC-Bench considers the full set of entities instantiated in generated videos, covering both prompt-specified and unspecified entities.
This enables us to assess two complementary requirements of coherent multi-shot generation: generated content should faithfully satisfy explicit textual constraints, while entities instantiated in the visual world should remain coherent within and across shots.

To implement this evaluation, we develop an agent-based pipeline that first discovers entities directly from generated videos with an object detection model~\cite{fu2025wedetect} and determines their correspondence to the prompts.
Based on this entity set, UEC-Bench evaluates three complementary dimensions: \emph{prompt alignment}, which measures whether prompt-specified entities are faithfully realized; \emph{intra-shot consistency}, which evaluates the coherence of entities and their properties within individual shots; and \emph{inter-shot consistency}, which evaluates whether entities remain coherent across shots.
Finally, the three dimensions are aggregated into a unified score, enabling comprehensive evaluation of entity consistency within and across shots.

We apply UEC-Bench to a diverse set of video generation models.
UEC-Bench exhibits strong agreement with human judgments, validating its effectiveness in assessing entity consistency.
Our evaluation further reveals substantial consistency failures that are largely overlooked by existing metrics, particularly for entities not explicitly specified in the prompts.
These findings highlight the importance of evaluating not only how faithfully models follow textual instructions, but also how consistently they maintain visual content beyond explicit prompt specifications.

Our contributions are summarized as follows:
\begin{itemize}
\item We formulate \emph{unspecified entity consistency} as an automatically measurable dimension of multi-shot video generation, which earlier benchmarks assess only through holistic human judgment.
\item We introduce UEC-Bench, a benchmark that jointly evaluates prompt alignment, intra-shot consistency, and inter-shot consistency over the entities instantiated in generated multi-shot videos.
\item We conduct a systematic evaluation of diverse multi-shot video generation models, revealing consistency challenges in unspecified visual content that are overlooked by prompt-centric evaluation.
\end{itemize}

\section{Related Work}
\label{sec:related}

\paragraph{Multi-shot video generation.}
Current multi-shot video generation methods mainly follow two approaches.
One-pass methods~\cite{wu2026cinetrans,jia2026moga,wang2026multishotmaster,meng2026holocine,guo2025long,kara2025shotadapter} generate all shots of an episode jointly, and many of them~\cite{wu2026cinetrans,wang2026multishotmaster,meng2026holocine} build on the video diffusion model Wan~\cite{wan2025wan}.
Autoregressive methods~\cite{an2026onestory,zhang2025storymem,luo2026shotstream,meng2026causalcine} generate an episode shot by shot and condition each new shot on context retained from earlier shots.
For example, StoryMem~\cite{zhang2025storymem} also builds on Wan, and ShotStream~\cite{luo2026shotstream} combines Wan with the causal distillation of CausVid~\cite{yin2025slow}, Self Forcing~\cite{huang2026self}, and the long-video model LongLive~\cite{yang2025longlive}.
These multi-shot works measure inter-shot consistency mainly with the global feature similarity between shots~\cite{wang2024internvid,radford2021learning,oquab2023dinov2}.
Such a single similarity value cannot show which entity changes in which shot, and thus where a model improves and where it still falls short.
Their own limitations, such as an object reverting to its earlier state after an action~\cite{meng2026holocine} or inconsistent character appearance~\cite{zhang2025storymem}, appear only in a few qualitative examples.
UEC-Bench instead detects and scores such issues in every generated video, including those of entities that the prompt never mentions.

\paragraph{Multi-shot benchmarks.}
As multi-shot generation has advanced, several dedicated benchmarks have been proposed, differing mainly in what they evaluate.
Story-level benchmarks such as MSVBench~\cite{shi2026msvbench}, MSAVBench~\cite{wei2026msavbench}, and DramaChain Bench~\cite{shi2026dramachain} score each video along predefined dimensions with large multimodal models, expert models, or agentic judges.
Subject- and entity-level benchmarks track selected content across shots: PersonaShot~\cite{wang2026personashot} evaluates the continuity of people, and GroundBench~\cite{lai2026groundshot} and EntityBench~\cite{he2026entitybench}, the latter closest to our work, evaluate the entities listed in scripts or per-shot schedules.
These benchmarks fix the evaluated content in advance, so entities that the model introduces on its own, such as background objects and accessories, are not evaluated, and they mainly report scores per dimension or entity rather than which problem occurs, in which shot, and to which entity.
UEC-Bench instead evaluates every entity found in the generated video, whether or not the prompt specifies it, and reports each inconsistency as a verified issue localized to shots, frames, and entities, from which the intra- and inter-shot scores are computed (Table~\ref{tab:related}).

\paragraph{Content beyond the prompt.}
Evaluations of video generation treat content the prompt leaves unspecified as a degree of freedom, an error, or something that must stay consistent.
DiVid~\cite{hu2026divid} measures how much samples of the same prompt vary in subject, scene, motion, and camera, while OPIS~\cite{wang2026opis} counts any object absent from the input image as a presence failure.
For consistency, StoryBench~\cite{bugliarello2023storybench} asks raters whether clothes or rooms change when the caption describes no change, World Consistency Score~\cite{rakheja2025world} penalizes objects that appear or disappear without cause, and Orbia~\cite{lu2026orbia} and WorldMemBench~\cite{zhou2026worldmembench} check whether content persists when the camera revisits it.
All of them evaluate a single continuous video, and none tracks model-added entities across cuts.
UEC-Bench adopts the consistency view for multi-shot video: content the model adds is legitimate, but once instantiated, it becomes an implicit requirement for the rest of the episode.
VMem-Bench~\cite{chen2026more} also finds that long-range recall can stem from the prompt re-describing an entity, which motivates evaluating separately the entities whose appearance the text never states.

\begin{table}[t]
\caption{\textbf{Comparison with Representative Benchmarks.}
Unsp.: entity-level evaluation of prompt unspecified entities; Verif.: independent re-examination of every finding; Issue: problems reported with type and location; Entity: judgments on individual entities; Intra/Inter: separate intra- and inter-shot consistency; Multi: protocol defined over shots.
\cmark: met; \pmark: partly met; \xmark: not met.}
\label{tab:related}
\centering
\small
\setlength{\tabcolsep}{2.5pt}
\begin{tabular}{@{}lcccccc@{}}
\toprule
Benchmark & Unsp. & Verif. & Issue & Entity & Intra/Inter & Multi \\
\midrule
VBench & \xmark & \xmark & \xmark & \pmark & \xmark & \xmark \\
MSVBench & \xmark & \xmark & \xmark & \pmark & \pmark & \cmark \\
EntityBench & \xmark & \xmark & \xmark & \cmark & \cmark & \cmark \\
PersonaShot & \xmark & \xmark & \xmark & \pmark & \cmark & \cmark \\
DramaChain & \xmark & \xmark & \cmark & \pmark & \cmark & \cmark \\
\midrule
UEC-Bench (ours) & \cmark & \cmark & \cmark & \cmark & \cmark & \cmark \\
\bottomrule
\end{tabular}
\end{table}

\section{UEC-Bench}
\label{sec:method}

\subsection{Overview}
\label{sec:overview}

\begin{figure}[t]
\centering
\includegraphics[width=\linewidth]{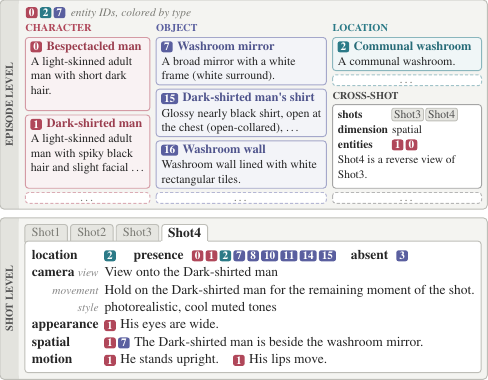}
\setlength{\abovecaptionskip}{3pt}
\caption{\textbf{Excerpt of the contract for a four-shot episode with 17 entities.}
The episode level describes each entity once and lists cross-shot requirements; the shot level has one block per shot that lists what the shot requires.}
\label{fig:contract}
\vspace{-1.2em}
\end{figure}

\paragraph{Benchmark and evaluation protocol.}
UEC-Bench consists of a set of episodes $\mathcal{D}=\{(P_i,C_i)\}_{i=1}^{|\mathcal{D}|}$, each pairing a prompt $P_i$ with a contract $C_i$.
The contract (Section~\ref{sec:construction}) organizes the requirements derived from the prompt at both episode and shot levels.
At the episode level, it describes each entity the prompt specifies once for all shots and lists the requirements that compare two or more shots.
At the shot level, it records the camera setup, any stated visual style, and the entities expected to be present, together with the requirements on their appearance, spatial relations, and motion (Figure~\ref{fig:contract}).
Given a model $M$ under test, we feed each prompt to the model and obtain a multi-shot video $V_i=M(P_i)=(v_i^1,\dots,v_i^{N_i})$, where $v_i^n$ is its $n$-th shot.
The evaluator takes $(P_i,C_i,V_i)$ and returns a set of verified issues $\mathcal{I}_i=\mathcal{I}_i^{\mathrm{VP}}\cup\mathcal{I}_i^{\mathrm{VV}}$, where $\mathcal{I}_i^{\mathrm{VP}}$ holds the video-prompt alignment issues, in which the video contradicts the prompt, and $\mathcal{I}_i^{\mathrm{VV}}$ the intra-video alignment issues, in which content of the video conflicts with other content of the same video; the latter are divided into intra-shot and inter-shot issues, $\mathcal{I}_i^{\mathrm{VV}}=\mathcal{I}_i^{\mathrm{intra}}\cup\mathcal{I}_i^{\mathrm{inter}}$.
Each of the three subsets $\mathcal{I}_i^{d}$, $d\in\{\mathrm{VP},\mathrm{intra},\mathrm{inter}\}$, is converted into an episode score $s_i^{d}$ (Section~\ref{sec:scoring}), namely the prompt alignment, intra-shot consistency, and inter-shot consistency scores, and the score of $M$ on dimension $d$ is the average $S^{d}(M)=\frac{1}{|\mathcal{D}|}\sum_{i} s_i^{d}$.

\paragraph{Specified and unspecified entities.}
An \emph{entity} is a character, location, or object that appears in a generated video.
Let $\mathcal{E}(V_i)$ denote the set of entities instantiated in $V_i$.
We partition this set into specified entities $\mathcal{E}_i^{\mathrm{s}}$ and unspecified entities $\mathcal{E}_i^{\mathrm{u}}$, $\mathcal{E}(V_i)=\mathcal{E}_i^{\mathrm{s}}\cup\mathcal{E}_i^{\mathrm{u}}$, depending on whether they are listed in the contract.
Unspecified entities may include background objects, unscripted bystanders, worn items, and scene elements the text does not mention.
As they exist only in the generated video, they must be discovered from it (Section~\ref{sec:discovery}).
Each issue records the shots and frames that show it and its \emph{subject entities}, those whose identity, appearance, state, presence, or position is affected.

\paragraph{Evaluation dimensions.}
\label{sec:dimensions}
The two kinds of issues defined above differ in their reference: the contract for video-prompt alignment ($\mathcal{I}_i^{\mathrm{VP}}$) and the video itself for intra-video alignment ($\mathcal{I}_i^{\mathrm{VV}}$).
An issue in $\mathcal{I}_i^{\mathrm{VP}}$ contradicts a requirement of $C_i$, such as an entity description, a spatial relation, or a transition between two shots. Each shot is compared with the contract text alone without considering other shots.

An issue in $\mathcal{I}_i^{\mathrm{VV}}$ is a change of the same entity or relation between entities that nothing explains.
The prompt only decides whether a change is permissible: a change the prompt calls for or allows, such as a described action, is not an issue, and neither is a change explained by camera motion, different viewpoints, or occlusion.
The issues are discovered by comparing visual information in different parts of the video.
An issue is \emph{intra-shot} ($\mathcal{I}_i^{\mathrm{intra}}$) when the conflicting come from one shot and \emph{inter-shot} ($\mathcal{I}_i^{\mathrm{inter}}$) when they come from different shots.

The two kinds of issues are not exclusive: a described attribute that changes between shots is an intra-video issue and, in the shots that violate the text, also a video-prompt issue, each counted in its own dimension (Table~\ref{tab:boundary}).

\begin{table}[t]
\caption{\textbf{Assignment of typical discrepancies.}
VP: video-prompt alignment; VV: intra-video alignment; $^{*}$only for the shots that violate the text.}
\label{tab:boundary}
\centering
\small
\begin{tabular}{@{}p{0.79\linewidth}cc@{}}
\toprule
Discrepancy & VP & VV \\
\midrule
A described attribute is wrong in every shot & \cmark & \\
An undescribed attribute changes between shots & & \cmark \\
A described attribute changes between shots and violates the text in some of them & \cmark$^{*}$ & \cmark \\
A shot satisfies the text but differs from another shot & & \cmark \\
A shot boundary shows no clear cut & \cmark & \\
\bottomrule
\end{tabular}
\par\vspace{1.2em}
\caption{\textbf{Benchmark statistics.}}
\label{tab:stats}
\setlength{\tabcolsep}{3pt}
\begin{tabular*}{\linewidth}{@{\extracolsep{\fill}}ll@{}}
\toprule
Episodes (SF20K / MuSS) & 200 (100 / 100) \\
Episodes with 4 / 6 / 8 shots & 120 / 60 / 20 \\
Shots (target duration) & 1,000 (5,000\,s) \\
Required transitions (all cuts) & 800 \\
Source works & 152 \\
Live action / animation / docum. & 173 / 24 / 3 \\
Words per shot prompt & 90.0 mean, 48--151 \\
\midrule
Entities & 4,438 \\
\quad character / location / object & 635 / 280 / 3,523 \\
\quad per episode & 22 median, 6--47 \\
Requirements: presence & 9,224 \\
\quad spatial / motion & 7,422 / 3,545 \\
\quad per-shot appearance & 1,550 \\
\quad cross-shot / absence & 92 / 8 \\
\bottomrule
\end{tabular*}
\vspace{-0.6em}
\end{table}

\subsection{Benchmark Construction}
\label{sec:construction}

\begin{figure*}[t]
\centering
\includegraphics[width=\textwidth]{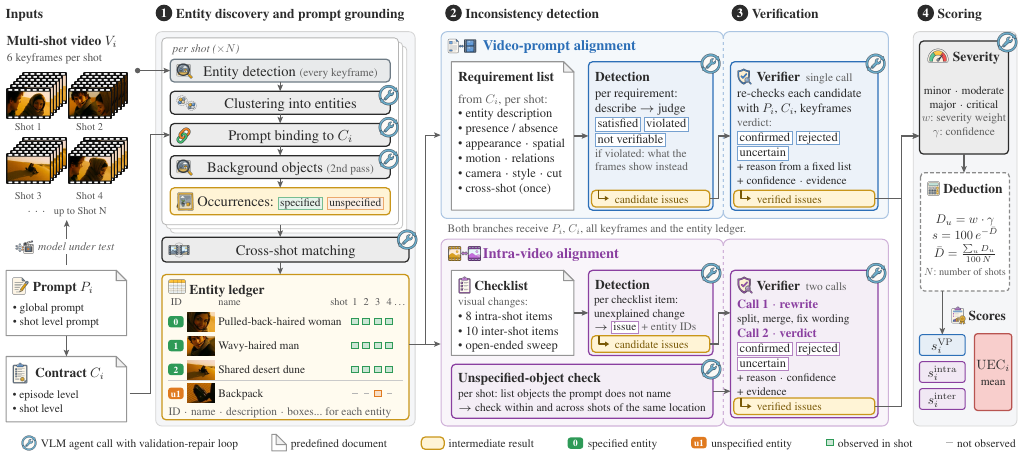}
\caption{\textbf{Overview of the evaluation pipeline.}
The evaluator builds an entity ledger, detects candidate issues for video-prompt and intra-video alignment in two branches, verifies each candidate, and turns the confirmed and uncertain issues into scores.}
\label{fig:pipeline}
\end{figure*}

\paragraph{Episodes.}
Episodes are windows of consecutive shots sampled from the SF20K~\cite{ghermi2026long} and MuSS~\cite{zhang2026muss} video indices, so that they inherit the shot structure, scene recurrence, and entity density of real narrative media.
For each window, GPT-6 Astra views about five frames per shot, records the visible facts with frame references, and writes a global prompt and one prompt per shot, which gives the framing or camera movement where the source shows it.
Table~\ref{tab:stats} summarizes the statistics.

\paragraph{Contracts.}
A contract is compiled from the requirements that the prompt states unambiguously, at the episode and shot levels described in Section~\ref{sec:overview}.
Each entity has an integer ID, a type (character, location, or object), and one description that gathers every stable attribute the prompt states; each shot block also records the main location of the shot (Figure~\ref{fig:contract}).
All contracts share one entity granularity so that episodes are comparable: an item that can be identified and followed on its own is an entity, a group seen as a whole is a single entity, and only entities the prompt names enter the contract, each expected only in the shots the prompt assigns it to.
The same principles define the entities in the generated video (Section~\ref{sec:discovery}), where those the prompt does not name are unspecified; the supplementary material gives the full rules.

\subsection{Evaluation Pipeline}
\label{sec:pipeline}

The evaluator works in four stages: \emph{Discovery}, \emph{Detection}, \emph{Verification} and \emph{Scoring} (Figure~\ref{fig:pipeline}).
\emph{Discovery} finds the entities in the video and divides them into specified and unspecified entities according to the contract (Section~\ref{sec:discovery}).
\emph{Detection} proposes candidate issues separately for video-prompt alignment and intra-video alignment (Section~\ref{sec:detection}).
\emph{Verification} re-examines each candidate and marks it as confirmed, rejected or uncertain (Section~\ref{sec:verification}).
\emph{Scoring} turns the confirmed and uncertain issues into a set of scores (Section~\ref{sec:scoring}).

Every stage uses VLM agents, and each agent output passes through a validation-repair loop.
A program checks the output against its schema and a set of deterministic rules, for example that every contract requirement is answered exactly once.
If the check fails, the agent receives the violations and corrects the output accordingly, repeating until the output passes or a fixed number of rounds is reached.

\subsubsection{Entity Discovery and Prompt Grounding}
\label{sec:discovery}

This stage builds an episode-level ledger of the entities in $V_i$ and divides them into specified and unspecified entities according to the contract.
Entities are detected and clustered within each shot, bound to contract entities, and then linked across shots.

Each shot is represented by six keyframes sampled uniformly from its start to its end.
We use an open-vocabulary detector~\cite{fu2025wedetect} to propose candidate boxes for entity occurrences on every keyframe.
A separate agent groups the occurrences into named entities and rejects false detections separately for each shot.
Whether a detection forms an entity of its own is decided by the granularity rules of the contracts (Section~\ref{sec:construction}).

Each shot-level entity is then bound to a contract entity, or left unbound, as decided by an agent.
For every keyframe in which a contract entity expected in the shot has no box, a text-guided detector~\cite{liu2024grounding} searches for it, and the highest-scoring box in that keyframe is attached to the entity. A keyframe that still has no box counts as unobserved.
This binding process happens within each shot before cross-shot linking, so that the identity of a specified entity comes from the contract rather than from visual similarity; otherwise a changed rendering could split one entity in two and hide the inconsistency the benchmark is designed to detect.
To improve the detection of small and background objects, which are easily missed, we add a background discovery step: an agent lists the visible objects that no entity covers, the text-guided detector searches for them, and the new occurrences are clustered and bound together with the existing ones.

After the shot-level discovery and binding, we match the entities across shots to form a complete episode-level ledger. Specified entities are linked through their contract IDs, and unspecified entities with no such IDs are linked by a matching agent. Given the keyframes and the prompts, the agent decides which of the occurrences show the same entity, or a specified entity whose binding was missed in its shot. It judges identity from appearance together with the relations of each occurrence, so that an entity rendered inconsistently across shots is not mistakenly split.

The output of this stage is an entity ledger containing all specified and unspecified entities discovered. Each record gives the entity's ID, name, a short description, and its occurrences, with the shot and keyframe indices and the bounding boxes. This ledger serves as reference information for the following inconsistency detection stage.

\subsubsection{Inconsistency Detection}
\label{sec:detection}

Based on the discovery results, this stage proposes candidate inconsistency issues of the generated multi-shot video.
The issues are classified into video-prompt alignment issues and intra-video alignment issues, as defined in Section~\ref{sec:dimensions}.
We detect these two types of issues separately.
Both receive all keyframes together with the prompt, the contract, and the entity ledger.

\paragraph{Video-prompt alignment.}
We use an agent to check a set of requirements one by one and report each violation as a candidate issue.
The requirements are derived deterministically from the contract and fall into a few groups: the presence, absence, and description of entities, the appearance, spatial, and motion requirements of each shot, the camera and style, the cuts between shots, and the requirements that span several shots.
For each requirement, the agent first describes what the frames show and then judges the requirement as satisfied, violated, or not verifiable.
Each issue quotes the violated requirement and states what the frames show instead.
The supplementary material gives the full requirement list and the answering rules.

\paragraph{Intra-video alignment.}
The intra-video alignment agent inspects the video with a predefined checklist, whose items name the aspects to compare.
It has eight intra-shot items, such as camera movement, clothing and background objects, and ten inter-shot items, such as faces, settings and spatial layout; the supplementary material lists all items.
To catch problems outside the checklist, it ends with an open-ended sweep, in which the agent re-examines all frames for any other visible inconsistency.

The agent checks one checklist item per call.
In each call, it looks up the entities that the item concerns in the entity ledger and compares them in the keyframes as the item specifies.
Since camera motion or an action in the prompt also changes what the frames show, a difference becomes an issue only if nothing explains it.

To strengthen the check on unspecified entities, which is our central goal, we further add an unspecified-object check.
In each shot, an agent lists the visible objects that the prompt does not name.
Another agent then checks each object within its shot and against the other shots of the same location, and adds the inconsistencies it finds to the checklist candidates for verification.

\subsubsection{Verification}
\label{sec:verification}

As a detection agent can report wrong issues, we use a verifier agent to re-examine every candidate, separately for the two kinds of issues.
Repeated reports of one problem are merged, so that the problem deducts only once.
A video-prompt alignment candidate already states one failed requirement and goes directly to the verdict.
An intra-video alignment candidate is first rewritten in a separate call to prevent wrong verdicts due to bundled problems or unclear wording.
For both kinds, the agent then judges each issue as confirmed, rejected, or uncertain, with a reason chosen from a fixed list, supporting evidence, and a confidence of high, medium, or low.

\subsubsection{Scoring and Aggregation}
\label{sec:scoring}

The scoring agent rates each confirmed or uncertain issue as minor, moderate, major, or critical from the verifier's text alone.
An issue $u$ deducts $D_u=w(\mathrm{sev}_u)\,\gamma(\mathrm{conf}_u)$, where $w$ is 10, 25, 50, or 80 for the four severities, and $\gamma$ is 1, 0.75, or 0.5 for high, medium, or low verifier confidence and 0.25 for an uncertain issue.
With $\mathcal{I}_i^{d}$ the confirmed and uncertain issues of dimension $d$ and $N_i$ the number of shots, the episode score is $s_i^{d}=100\exp\bigl(-\sum_{u\in\mathcal{I}_i^{d}}D_u/(100\,N_i)\bigr)$, which normalizes the deduction per shot, so that a longer episode does not score lower for its length alone.
The final UEC score is the mean of the three dimensions, $\mathrm{UEC}_i=\frac{1}{3}(s_i^{\mathrm{VP}}+s_i^{\mathrm{intra}}+s_i^{\mathrm{inter}})$, with all scores in $[0,100]$.

\section{Experiments}
\label{sec:exp}

\begin{table*}[t]
\caption{\textbf{Evaluation of Current Models on UEC-Bench.}
Top: baselines not designed for multi-shot generation; middle and bottom: one-pass and autoregressive multi-shot methods.
Unspecified share: unspecified entities among all entities, and confirmed intra-video issues with an unspecified subject entity among all confirmed intra-video issues.
Entities with issues: share of specified and of unspecified entities that are a subject of at least one confirmed intra-video issue.}
\label{tab:results}
\centering
\small
\begin{tabular*}{\textwidth}{@{\extracolsep{\fill}}lccc>{\columncolor{gray!12}}c@{\hspace{3em}}c@{\hspace{3em}}cccc@{}}
\toprule
\multirow{2}{*}{Model} & \multirow{2}{*}{Video-prompt} & \multicolumn{2}{c}{Intra-video} & \textbf{UEC}\rule[-1.1ex]{0pt}{0pt} & & \multicolumn{2}{c}{Unspecified share} & \multicolumn{2}{c}{Entities with issues} \\
\noalign{\vskip-\aboverulesep}
\cmidrule(lr){3-4}\cmidrule(lr){7-8}\cmidrule(l){9-10}
\noalign{\vskip-\belowrulesep}
 & & Intra-shot\rule{0pt}{2.6ex} & Inter-shot & (avg.) & & Entities & Issues & Specified & Unspecified \\
\cmidrule[\lightrulewidth](r){1-5}\cmidrule[\lightrulewidth](l){6-10}
Wan2.1-T2V-1.3B & 64.2 & 96.3 & 71.4 & 77.3 & & 32.1\% & 24.3\% & 19.6\% & 16.0\% \\
Wan2.2-T2V-A14B & 82.6 & 98.0 & 75.7 & 85.4 & & 35.7\% & 27.6\% & 15.4\% & 13.7\% \\
LongLive-1.3B & 37.3 & 87.1 & 88.6 & 71.0 & & 28.4\% & 32.7\% & 11.6\% & 15.3\% \\
\cmidrule[\lightrulewidth](r){1-5}\cmidrule[\lightrulewidth](l){6-10}
CineTrans-DiT-1.3B & 21.7 & 95.0 & 85.5 & 67.4 & & 17.8\% & 26.6\% & 8.7\% & 20.7\% \\
MultiShotMaster-1.3B & 55.1 & 97.4 & 90.4 & 81.0 & & 34.6\% & 28.3\% & 6.7\% & 7.0\% \\
HoloCine-A14B & 65.7 & 96.8 & \textbf{96.4} & 86.3 & & 24.3\% & 25.1\% & 3.4\% & 5.1\% \\
\cmidrule[\lightrulewidth](r){1-5}\cmidrule[\lightrulewidth](l){6-10}
ShotStream-1.3B & 44.7 & \textbf{98.7} & 90.6 & 78.0 & & 22.5\% & 34.9\% & 7.3\% & 12.7\% \\
StoryMem-A14B & \textbf{84.0} & 97.1 & 92.4 & \textbf{91.2} & & 31.8\% & 38.2\% & 5.5\% & 8.7\% \\
\bottomrule
\end{tabular*}
\end{table*}

We use UEC-Bench to evaluate eight video generation models and examine where they fail (Section~\ref{sec:exp-results}), followed by an analysis of its agreement with human judgment and advantage over direct VLM rating (Section~\ref{sec:exp-analysis}).
These analyses show that UEC-Bench can detect problems related to unspecified entities which the prompt-centric methods overlook, while agreeing with human judgments and remaining stable across repeated runs.

\subsection{Setup}
\label{sec:exp-setup}

\paragraph{Models.}
We evaluate five multi-shot methods and three baselines not designed for multi-shot generation: Wan2.1, Wan2.2~\cite{wan2025wan}, and LongLive~\cite{yang2025longlive}.
Among the multi-shot methods, CineTrans-DiT~\cite{wu2026cinetrans}, MultiShotMaster~\cite{wang2026multishotmaster}, and HoloCine~\cite{meng2026holocine} generate the whole episode in one pass, while ShotStream~\cite{luo2026shotstream} and StoryMem~\cite{zhang2025storymem} generate it shot by shot.
Each multi-shot method starts from the weights of Wan2.1 (CineTrans-DiT, MultiShotMaster, and ShotStream) or Wan2.2 (HoloCine and StoryMem), which we call its base model.
All videos are 832$\times$480 at 16 fps (MultiShotMaster at 15 fps) with about five seconds per shot.
Each model receives the shared prompts, adapted only as its input format requires.

\paragraph{Evaluator.}
All eight models are evaluated by the same pipeline (Section~\ref{sec:pipeline}) under one fixed configuration.
Every vision-language model call in it uses GPT-6 Luna at low reasoning effort.
Further details are provided in the supplementary material.

\subsection{Evaluation of Current Models}
\label{sec:exp-results}

\paragraph{Overall results.}
As shown in Table~\ref{tab:results}, all models perform well on intra-shot consistency, and they differ mainly in prompt alignment and inter-shot consistency.
On these two dimensions, larger models perform better: within each group, the A14B model has higher video-prompt and UEC scores than the 1.3B ones.
Most multi-shot methods reach higher inter-shot scores than their base models but follow the prompts less well; only StoryMem improves on both and has the best video-prompt and UEC scores.
LongLive and CineTrans-DiT have the lowest UEC scores, for reasons we analyze below.

\paragraph{Unspecified entities.}
Over all 1,600 videos, unspecified entities make up 28.9\% of all entities and are the subject of 28.8\% of the confirmed intra-video issues, mostly inter-shot changes in clothing, accessories and background objects that the prompts do not mention.
Relative to its base model, every multi-shot method reduces inter-shot issues of specified entities more than those of unspecified ones; StoryMem, for example, reduces them by 74\% and 51\%, respectively.
Consequently, in every multi-shot method an unspecified entity is more likely than a specified one to be the subject of a confirmed issue, whereas the opposite holds for the Wan baselines.
The multi-shot designs thus strengthen consistency mainly for prompted entities, and unspecified entities remain a source of inconsistency even for the best model.
Evaluations that check only prompted entities miss this gap; UEC-Bench reveals it by tracking every entity in the generated video, prompted or not, and attributing each verified issue to its subject entities.

\paragraph{Failure types.}
The Wan models fail mainly in inter-shot consistency as they generate shots independently.
Although StoryMem reduces inter-shot issues relative to its base model without losing prompt alignment,
the other multi-shot methods often leave out or miscount required entities, several times as often as their base models, leading to lower prompt alignment scores.
These issues per shot grow with the number of shots for the one-pass methods but not for the autoregressive ShotStream.
The two models with the lowest UEC fail in specific ways.
CineTrans-DiT often generates nearly static shots without the required actions, likely because our episodes far exceed the few-second clips it is designed for~\cite{wu2026cinetrans}.
LongLive cannot cut: at each prompt switch it keeps the view and gradually turns existing entities into those of the next prompt, such as a dog into a man, so required cuts are missing and identities drift within shots.
UEC-Bench reports these as confirmed issues, showing that its scores capture real failures in the videos.

\begin{figure*}[t]
\centering
\includegraphics[width=\textwidth]{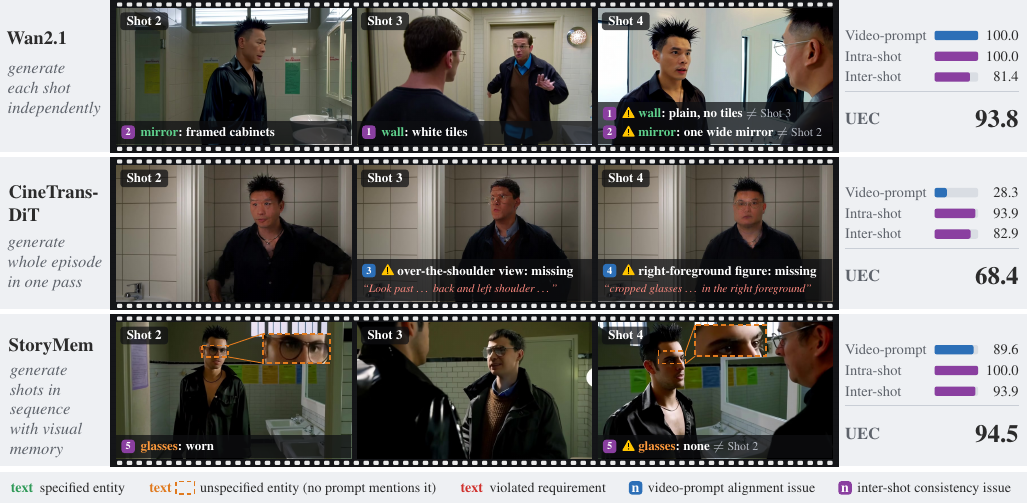}
\caption{\textbf{Issues confirmed by UEC-Bench for three models on one episode.}
In episode \#101 (Shots 2--4), Wan2.1 changes the washroom wall and mirror across shots (1, 2), CineTrans-DiT misses the over-the-shoulder view of Shot 3 and the cropped right-foreground figure of Shot 4 that the prompts require (3, 4), and StoryMem changes the glasses of the dark-shirted man, which no prompt mentions (5).
Each row ends with the scores of its video.}
\label{fig:qualitative}
\end{figure*}

\paragraph{Qualitative examples.}
Figure~\ref{fig:qualitative} shows one episode generated by three models with different strategies.
Wan2.1 changes the washroom wall and mirror between shots, and CineTrans-DiT ignores the framing the prompts require in Shots 3 and 4, while StoryMem keeps both men and the washroom consistent.
Even for StoryMem, the dark-shirted man wears glasses in Shot 2 but not in Shot 4. No prompt mentions these glasses, so a prompt-centric evaluation does not check them, whereas UEC-Bench reports the change as an inter-shot issue of an unspecified entity.

\subsection{Benchmark Analysis}
\label{sec:exp-analysis}
\label{sec:exp-human}

To examine whether UEC-Bench agrees with human judgment, we sample 24 episodes and take the videos five models (Wan2.1/2.2, CineTrans-DiT, HoloCine, StoryMem) generate for them.
Ten raters score these 120 videos from 1 (worst) to 5 (best) for consistency, five raters per video.
Agreement with the mean human rating is measured in two ways: Spearman's $\rho$ over all videos reflects how well a metric ranks videos, and Kendall's $\tau_b$ within each episode, averaged over episodes, reflects how well it ranks models on the same prompt.
Baselines are ViCLIP similarity~\cite{wang2024internvid}, VBench consistency~\cite{huang2024vbench}, and a direct VLM rating.

As shown in Table~\ref{tab:human}, UEC-Bench agrees with the human ratings on both measures.
The agreement is not tied to one evaluation model or reasoning effort, as GPT-6 Luna and Gemini~3.8 Flash at medium effort give values similar to those of GPT-6 Luna at low effort.
ViCLIP similarity and VBench consistency correlate only weakly or even negatively with the raters.
Although the direct VLM rating comes closer, UEC-Bench still stays ahead on both measures.
What's more, compared with a direct rating that only gives scores and short reasons, UEC-Bench answers 165 contract requirements per video and re-examines those left unjudged through its repair loop.
With clear scoring criteria, UEC-Bench can separate videos more finely: a direct rating has only a few score levels and often gives the same score to videos of the same prompt that the raters rate differently, while UEC-Bench does not.
Each issue names its shots, frames, and entities (Figure~\ref{fig:qualitative}), so score losses can be traced to failure types and entities (Section~\ref{sec:exp-results}), unlike a direct rating.

\begin{table}[t]
\caption{\textbf{Correlation with the overall human rating.}
Spearman's $\rho$ over all videos and Kendall's $\tau_b$ within episodes.}
\label{tab:human}
\centering
\footnotesize
\setlength{\tabcolsep}{2.6pt}
\begin{tabular}{@{}lcccccc@{}}
\toprule
\multirow{3}{*}{Method} & \multicolumn{3}{c}{UEC-Bench} & \multirow{3}{*}{VBench} & \multirow{3}{*}{ViCLIP} & Direct VLM \\
\cmidrule(lr){2-4}\cmidrule(l){7-7}
 & Luna & Luna & Gemini & & & Luna \\
 & low & medium & medium & & & low \\
\midrule
Spearman's $\rho$ & 0.75 & \textbf{0.76} & 0.75 & $-$0.25 & 0.15 & 0.68 \\
Kendall's $\tau_b$ & \textbf{0.68} & 0.67 & 0.64 & $-$0.38 & 0.17 & 0.49 \\
\bottomrule
\end{tabular}
\end{table}

\section{Conclusion}
\label{sec:conclusion}

We introduced UEC-Bench, a benchmark that evaluates multi-shot video generation over the entities a model actually instantiates, whether or not the prompt mentions them.
Its agent-based pipeline discovers specified and unspecified entities, checks prompt alignment against contracts derived from the prompts and intra- and inter-shot consistency against a checklist, and verifies every detected issue before scoring.
Across the eight evaluated models, unspecified entities are the subject of 28.8\% of the confirmed intra-video issues, and multi-shot designs improve consistency mainly for prompted entities, a gap that prompt-centric evaluation cannot reveal.
UEC-Bench agrees with human ratings both in ranking videos and in ranking models on the same prompt, without depending on a specific evaluation model or reasoning effort.
It also agrees better with the raters than a direct VLM rating, and its scores rest on requirement-level checks and verified issues located in shots, frames, and entities, so a low score can be explained, not only observed.

\paragraph{Limitations.}
Entity discovery is influenced by the recall of the detector, and our setting of six keyframes per shot can miss brief artifacts.
Without 3D information, the VLM agents judge 3D spatial consistency less reliably.

\section*{LLM Usage Statement}
We used LLM-based assistants (Claude Code and Codex) to write and debug the evaluation and analysis code, to survey related work, and to revise the wording and grammar of this paper and typeset its figures and tables.
VLMs are also part of the benchmark (Section~\ref{sec:method}): GPT-6 Astra writes the prompts, Claude Sonnet compiles the contracts, and GPT-6 Luna and Gemini~3.8 Flash serve as the evaluation agents and the direct-rating baseline.
The authors reviewed and verified all LLM-assisted \mbox{code, analyses, and text.}

{
    \small
    \bibliographystyle{ieeenat_fullname}
    \bibliography{main}

\begin{thebibliography}{42}
\providecommand{\natexlab}[1]{#1}
\providecommand{\url}[1]{\texttt{#1}}
\expandafter\ifx\csname urlstyle\endcsname\relax
  \providecommand{\doi}[1]{doi: #1}\else
  \providecommand{\doi}{doi: \begingroup \urlstyle{rm}\Url}\fi

\bibitem[An et~al.(2026)An, Jia, Qiu, Zhou, Huang, Liu, Ren, Kahatapitiya, Liu,
  He, et~al.]{an2026onestory}
Zhaochong An, Menglin Jia, Haonan Qiu, Zijian Zhou, Xiaoke Huang, Zhiheng Liu,
  Weiming Ren, Kumara Kahatapitiya, Ding Liu, Sen He, et~al.
\newblock Onestory: Coherent multi-shot video generation with adaptive memory.
\newblock In \emph{Proceedings of the IEEE/CVF conference on computer vision
  and pattern recognition}, pages 16173--16184, 2026.

\bibitem[Blattmann et~al.(2023)Blattmann, Dockhorn, Kulal, Mendelevitch,
  Kilian, Lorenz, Levi, English, Voleti, Letts, et~al.]{stablevideodiffusion}
Andreas Blattmann, Tim Dockhorn, Sumith Kulal, Daniel Mendelevitch, Maciej
  Kilian, Dominik Lorenz, Yam Levi, Zion English, Vikram Voleti, Adam Letts,
  et~al.
\newblock Stable video diffusion: Scaling latent video diffusion models to
  large datasets.
\newblock \emph{arXiv preprint arXiv:2311.15127}, 2023.

\bibitem[Bugliarello et~al.(2023)Bugliarello, Moraldo, Villegas, Babaeizadeh,
  Saffar, Zhang, Erhan, Ferrari, Kindermans, and
  Voigtlaender]{bugliarello2023storybench}
Emanuele Bugliarello, H~Hernan Moraldo, Ruben Villegas, Mohammad Babaeizadeh,
  Mohammad~Taghi Saffar, Han Zhang, Dumitru Erhan, Vittorio Ferrari, Pieter-Jan
  Kindermans, and Paul Voigtlaender.
\newblock Storybench: A multifaceted benchmark for continuous story
  visualization.
\newblock \emph{Advances in Neural Information Processing Systems},
  36:\penalty0 78095--78125, 2023.

\bibitem[Chen et~al.(2026)Chen, Shi, Wang, Xu, Wang, Hong, Zhang, Ma, and
  Zhang]{chen2026more}
Yuzhuo Chen, Huafeng Shi, Xinyu Wang, Yulong Xu, Yucheng Wang, Haoqin Hong,
  Guoxin Zhang, Zehua Ma, and Weiming Zhang.
\newblock More memory, better videos? a controlled study of visual memory in
  long-video generation.
\newblock \emph{arXiv preprint arXiv:2610.35001}, 2026.

\bibitem[Fu et~al.(2025)Fu, Su, Rao, LYU, Xie, and Zheng]{fu2025wedetect}
Shenghao Fu, Yukun Su, Fengyun Rao, Jing LYU, Xiaohua Xie, and Wei-Shi Zheng.
\newblock Wedetect: Fast open-vocabulary object detection as retrieval.
\newblock \emph{arXiv preprint arXiv:2512.12309}, 2025.

\bibitem[Ghermi et~al.(2026)Ghermi, Wang, Kalogeiton, and
  Laptev]{ghermi2026long}
Ridouane Ghermi, Xi Wang, Vicky Kalogeiton, and Ivan Laptev.
\newblock Long story short: Story-level video understanding from 20k short
  films: R. ghermi et al.
\newblock \emph{International Journal of Computer Vision}, 134\penalty0
  (8):\penalty0 396, 2026.

\bibitem[Guo et~al.(2025)Guo, Yang, Yang, Ma, Lin, Yang, Lin, and
  Jiang]{guo2025long}
Yuwei Guo, Ceyuan Yang, Ziyan Yang, Zhibei Ma, Zhijie Lin, Zhenheng Yang, Dahua
  Lin, and Lu Jiang.
\newblock Long context tuning for video generation.
\newblock In \emph{Proceedings of the IEEE/CVF International Conference on
  Computer Vision}, pages 17281--17291, 2025.

\bibitem[He et~al.(2026)He, Wei, Yang, and Ordonez]{he2026entitybench}
Ruozhen He, Meng Wei, Ziyan Yang, and Vicente Ordonez.
\newblock Entitybench: Towards entity-consistent long-range multi-shot video
  generation.
\newblock \emph{arXiv preprint arXiv:2605.15199}, 2026.

\bibitem[Ho et~al.(2022)Ho, Salimans, Gritsenko, Chan, Norouzi, and Fleet]{vdm}
Jonathan Ho, Tim Salimans, Alexey Gritsenko, William Chan, Mohammad Norouzi,
  and David Fleet.
\newblock Video diffusion models.
\newblock \emph{Advances in neural information processing systems},
  35:\penalty0 8633--8646, 2022.

\bibitem[Hu et~al.(2026)Hu, Ren, Yang, Song, Wu, Ge, and Jin]{hu2026divid}
Huanran Hu, Zihui Ren, Dingyi Yang, Zhinan Song, Guozheng Wu, Tiezheng Ge, and
  Qin Jin.
\newblock Divid: Diagnosing dimension-specific diversity collapse in video
  generation models.
\newblock \emph{arXiv preprint arXiv:2610.01661}, 2026.

\bibitem[Huang et~al.(2026)Huang, Li, He, Zhou, and Shechtman]{huang2026self}
Xun Huang, Zhengqi Li, Guande He, Mingyuan Zhou, and Eli Shechtman.
\newblock Self forcing: Bridging the train-test gap in autoregressive video
  diffusion.
\newblock \emph{Advances in Neural Information Processing Systems},
  38:\penalty0 167283--167308, 2026.

\bibitem[Huang et~al.(2024)Huang, He, Yu, Zhang, Si, Jiang, Zhang, Wu, Jin,
  Chanpaisit, et~al.]{huang2024vbench}
Ziqi Huang, Yinan He, Jiashuo Yu, Fan Zhang, Chenyang Si, Yuming Jiang, Yuanhan
  Zhang, Tianxing Wu, Qingyang Jin, Nattapol Chanpaisit, et~al.
\newblock Vbench: Comprehensive benchmark suite for video generative models.
\newblock In \emph{2024 IEEE/CVF Conference on Computer Vision and Pattern
  Recognition (CVPR)}, pages 21807--21818. IEEE, 2024.

\bibitem[Jia et~al.(2026)Jia, Lu, Huang, Wang, Huang, Chen, Jiang, Mao,
  et~al.]{jia2026moga}
Weinan Jia, Yuning Lu, Mengqi Huang, Hualiang Wang, Binyuan Huang, Nan Chen,
  Jidong Jiang, Zhendong Mao, et~al.
\newblock Moga: Mixture-of-groups attention for end-to-end long video
  generation.
\newblock In \emph{International Conference on Learning Representations}, pages
  115944--115964, 2026.

\bibitem[Kara et~al.(2025)Kara, Singh, Liu, Ceylan, Rehg, and
  Hinz]{kara2025shotadapter}
Ozgur Kara, Krishna~Kumar Singh, Feng Liu, Duygu Ceylan, James~M Rehg, and
  Tobias Hinz.
\newblock Shotadapter: Text-to-multi-shot video generation with diffusion
  models.
\newblock In \emph{2025 IEEE/CVF Conference on Computer Vision and Pattern
  Recognition (CVPR)}, pages 28405--28415. IEEE, 2025.

\bibitem[Kong et~al.(2024)Kong, Tian, Zhang, Min, Dai, Zhou, Xiong, Li, Wu,
  Zhang, et~al.]{kong2024hunyuanvideo}
Weijie Kong, Qi Tian, Zijian Zhang, Rox Min, Zuozhuo Dai, Jin Zhou, Jiangfeng
  Xiong, Xin Li, Bo Wu, Jianwei Zhang, et~al.
\newblock Hunyuanvideo: A systematic framework for large video generative
  models.
\newblock \emph{arXiv preprint arXiv:2412.03603}, 2024.

\bibitem[Lai et~al.(2026)Lai, Shao, Dou, Zhu, and Wang]{lai2026groundshot}
Yixuan Lai, Tianjia Shao, Weijia Dou, Siyu Zhu, and Jingdong Wang.
\newblock Groundshot: Visually consistent multi-shot long video generation via
  entity-grounded shot scheduling.
\newblock \emph{arXiv preprint arXiv:2606.20799}, 2026.

\bibitem[Liu et~al.(2024{\natexlab{a}})Liu, Zeng, Ren, Li, Zhang, Yang, Jiang,
  Li, Yang, Su, et~al.]{liu2024grounding}
Shilong Liu, Zhaoyang Zeng, Tianhe Ren, Feng Li, Hao Zhang, Jie Yang, Qing
  Jiang, Chunyuan Li, Jianwei Yang, Hang Su, et~al.
\newblock Grounding dino: Marrying dino with grounded pre-training for open-set
  object detection.
\newblock In \emph{European conference on computer vision}, pages 38--55.
  Springer, 2024{\natexlab{a}}.

\bibitem[Liu et~al.(2024{\natexlab{b}})Liu, Cun, Liu, Wang, Zhang, Chen, Liu,
  Zeng, Chan, and Shan]{liu2024evalcrafter}
Yaofang Liu, Xiaodong Cun, Xuebo Liu, Xintao Wang, Yong Zhang, Haoxin Chen,
  Yang Liu, Tieyong Zeng, Raymond Chan, and Ying Shan.
\newblock Evalcrafter: Benchmarking and evaluating large video generation
  models.
\newblock In \emph{2024 IEEE/CVF Conference on Computer Vision and Pattern
  Recognition (CVPR)}, pages 22139--22149. IEEE, 2024{\natexlab{b}}.

\bibitem[Lu et~al.(2026)Lu, Fang, Lim, Wu, Wang, Chen, Qian, Liao, Chow, Kong,
  et~al.]{lu2026orbia}
Dongyue Lu, Guian Fang, Hur Lim, Yonggang Wu, Yanrong Wang, Hongshan Chen,
  Shengju Qian, Yue Liao, Wei Chow, Lingdong Kong, et~al.
\newblock Orbia: Do generated videos form a consistent 3d world?
\newblock \emph{arXiv preprint arXiv:2610.20693}, 2026.

\bibitem[Luo et~al.(2026)Luo, Shi, Zhuang, Chen, Liu, Wang, Wan, and
  Xue]{luo2026shotstream}
Yawen Luo, Xiaoyu Shi, Junhao Zhuang, Yutian Chen, Quande Liu, Xintao Wang,
  Pengfei Wan, and Tianfan Xue.
\newblock Shotstream: Streaming multi-shot video generation for interactive
  storytelling.
\newblock In \emph{European Conference on Computer Vision}, pages 193--209.
  Springer, 2026.

\bibitem[Meng et~al.(2026{\natexlab{a}})Meng, Liu, Ouyang, Wang, Cheng, Yu,
  Wang, Li, Zhu, Zeng, et~al.]{meng2026causalcine}
Yihao Meng, Zichen Liu, Hao Ouyang, Qiuyu Wang, Ka~Leong Cheng, Yue Yu, Hanlin
  Wang, Haobo Li, Jiapeng Zhu, Yanhong Zeng, et~al.
\newblock Causalcine: Real-time autoregressive generation for multi-shot video
  narratives.
\newblock \emph{arXiv preprint arXiv:2605.12496}, 2026{\natexlab{a}}.

\bibitem[Meng et~al.(2026{\natexlab{b}})Meng, Ouyang, Yu, Wang, Wang, Cheng,
  Wang, Ma, Li, Chen, et~al.]{meng2026holocine}
Yihao Meng, Hao Ouyang, Yue Yu, Qiuyu Wang, Wen Wang, Ka~Leong Cheng, Hanlin
  Wang, Shuailei Ma, Yixuan Li, Cheng Chen, et~al.
\newblock Holocine: Holistic generation of cinematic multi-shot long video
  narratives.
\newblock In \emph{Proceedings of the IEEE/CVF conference on computer vision
  and pattern recognition}, pages 461--471, 2026{\natexlab{b}}.

\bibitem[Oquab et~al.(2023)Oquab, Darcet, Moutakanni, Vo, Szafraniec, Khalidov,
  Fernandez, Haziza, Massa, El-Nouby, et~al.]{oquab2023dinov2}
Maxime Oquab, Timoth{\'e}e Darcet, Th{\'e}o Moutakanni, Huy Vo, Marc
  Szafraniec, Vasil Khalidov, Pierre Fernandez, Daniel Haziza, Francisco Massa,
  Alaaeldin El-Nouby, et~al.
\newblock Dinov2: Learning robust visual features without supervision.
\newblock \emph{arXiv preprint arXiv:2304.07193}, 2023.

\bibitem[Radford(2021)]{radford2021learning}
A Radford.
\newblock Learning transferable visual models from natural language
  supervision.
\newblock \emph{arXiv preprint arXiv:2103.00020}, 2021.

\bibitem[Rakheja et~al.(2025)Rakheja, Ashdhir, Bhattacharjee, and
  Sharma]{rakheja2025world}
Akshat Rakheja, Aarsh Ashdhir, Aryan Bhattacharjee, and Vanshika Sharma.
\newblock World consistency score: a unified metric for video generation
  quality.
\newblock \emph{arXiv preprint arXiv:2508.00144}, 2025.

\bibitem[Shi et~al.(2026{\natexlab{a}})Shi, Chen, Jiang, Chen, Sheng, Liu,
  Zhang, Wang, Lu, Lu, et~al.]{shi2026dramachain}
Haoyuan Shi, Mingtao Chen, Shuo Jiang, Ziyan Chen, Xuyi Sheng, Yiming Liu, Ying
  Zhang, Miao Wang, Jianxiang Lu, Fanyang Lu, et~al.
\newblock Dramachain bench: An end-to-end benchmark for short-drama generation.
\newblock \emph{arXiv preprint arXiv:2609.00646}, 2026{\natexlab{a}}.

\bibitem[Shi et~al.(2026{\natexlab{b}})Shi, Li, Deng, Xu, Chen, Wang, Hu, and
  Zhang]{shi2026msvbench}
Haoyuan Shi, Yunxin Li, Nanhao Deng, Zhenran Xu, Xinyu Chen, Longyue Wang,
  Baotian Hu, and Min Zhang.
\newblock Msvbench: Towards human-level evaluation of multi-shot video
  generation.
\newblock In \emph{Findings of the Association for Computational Linguistics:
  ACL 2026}, pages 24034--24058, 2026{\natexlab{b}}.

\bibitem[Singer et~al.(2022)Singer, Polyak, Hayes, Yin, An, Zhang, Hu, Yang,
  Ashual, Gafni, et~al.]{makeavideo}
Uriel Singer, Adam Polyak, Thomas Hayes, Xi Yin, Jie An, Songyang Zhang, Qiyuan
  Hu, Harry Yang, Oron Ashual, Oran Gafni, et~al.
\newblock Make-a-video: Text-to-video generation without text-video data.
\newblock \emph{arXiv preprint arXiv:2209.14792}, 2022.

\bibitem[Wan et~al.(2025)Wan, Wang, Ai, Wen, Mao, Xie, Chen, Yu, Zhao, Yang,
  et~al.]{wan2025wan}
Team Wan, Ang Wang, Baole Ai, Bin Wen, Chaojie Mao, Chen-Wei Xie, Di Chen,
  Feiwu Yu, Haiming Zhao, Jianxiao Yang, et~al.
\newblock Wan: Open and advanced large-scale video generative models.
\newblock \emph{arXiv preprint arXiv:2503.20314}, 2025.

\bibitem[Wang et~al.(2026{\natexlab{a}})Wang, Yu, Zhang, Wang, Jin, Zhou, Wang,
  Yi, Li, Wang, et~al.]{wang2026opis}
Hao Wang, Tao Yu, Liuzhou Zhang, HeXin Wang, Haopeng Jin, Yuxuan Zhou, Xinming
  Wang, Hongzhu Yi, Xinye Li, Yuanlei Wang, et~al.
\newblock Opis: An input-grounded benchmark for multi-object memory in video
  world models.
\newblock \emph{arXiv preprint arXiv:2609.35052}, 2026{\natexlab{a}}.

\bibitem[Wang et~al.(2026{\natexlab{b}})Wang, Shi, Li, Bian, Liu, Lu, Wang,
  Wan, Gai, and Jia]{wang2026multishotmaster}
Qinghe Wang, Xiaoyu Shi, Baolu Li, Weikang Bian, Quande Liu, Huchuan Lu, Xintao
  Wang, Pengfei Wan, Kun Gai, and Xu Jia.
\newblock Multishotmaster: A controllable multi-shot video generation
  framework.
\newblock In \emph{Proceedings of the IEEE/CVF Conference on Computer Vision
  and Pattern Recognition}, pages 16268--16278, 2026{\natexlab{b}}.

\bibitem[Wang et~al.(2024)Wang, He, Li, Li, Yu, Ma, Li, Chen, Chen, Wang,
  et~al.]{wang2024internvid}
Yi Wang, Yinan He, Yizhuo Li, Kunchang Li, Jiashuo Yu, Xin Ma, Xinhao Li, Guo
  Chen, Xinyuan Chen, Yaohui Wang, et~al.
\newblock Internvid: A large-scale video-text dataset for multimodal
  understanding and generation.
\newblock In \emph{International Conference on Learning Representations}, pages
  42055--42079, 2024.

\bibitem[Wang et~al.(2026{\natexlab{c}})Wang, Chen, Hu, Yi, Hong, Feng, Cao,
  Wang, Ma, and Zhang]{wang2026personashot}
Yuji Wang, Yuheng Chen, Teng Hu, Ran Yi, Yijia Hong, Han Feng, Weijian Cao,
  Chengjie Wang, Lizhuang Ma, and Jiangning Zhang.
\newblock Personashot: Benchmarking person-centric narrative continuity in
  multi-shot video generation.
\newblock \emph{arXiv preprint arXiv:2608.16717}, 2026{\natexlab{c}}.

\bibitem[Wei et~al.(2026)Wei, Han, Chen, Li, Jiang, Liu, Li, Qing, Wang, Xing,
  et~al.]{wei2026msavbench}
Yujie Wei, Yujin Han, Zhekai Chen, Yongming Li, Kaixun Jiang, Zhihang Liu,
  Quanhao Li, Zhiwu Qing, Xiang Wang, Zhen Xing, et~al.
\newblock Msavbench: Towards comprehensive and reliable evaluation of
  multi-shot audio-video generation.
\newblock \emph{arXiv preprint arXiv:2605.20183}, 2026.

\bibitem[Wu et~al.(2026)Wu, Gao, Qiao, Wang, and Chen]{wu2026cinetrans}
Xiaoxue Wu, Bingjie Gao, Yu Qiao, Yaohui Wang, and Xinyuan Chen.
\newblock Cinetrans: Learning to generate videos with cinematic transitions via
  masked diffusion models.
\newblock In \emph{International Conference on Learning Representations}, pages
  60198--60230, 2026.

\bibitem[Yang et~al.(2025{\natexlab{a}})Yang, Huang, Chu, Xiao, Zhao, Wang, Li,
  Xie, Chen, Lu, et~al.]{yang2025longlive}
Shuai Yang, Wei Huang, Ruihang Chu, Yicheng Xiao, Yuyang Zhao, Xianbang Wang,
  Muyang Li, Enze Xie, Yingcong Chen, Yao Lu, et~al.
\newblock Longlive: Real-time interactive long video generation.
\newblock \emph{arXiv preprint arXiv:2509.22622}, 2025{\natexlab{a}}.

\bibitem[Yang et~al.(2025{\natexlab{b}})Yang, Teng, Zheng, Ding, Huang, Xu,
  Yang, Hong, Zhang, Feng, et~al.]{yang2025cogvideox}
Zhuoyi Yang, Jiayan Teng, Wendi Zheng, Ming Ding, Shiyu Huang, Jiazheng Xu,
  Yuanming Yang, Wenyi Hong, Xiaohan Zhang, Guanyu Feng, et~al.
\newblock Cogvideox: Text-to-video diffusion models with an expert transformer.
\newblock In \emph{International Conference on Learning Representations}, pages
  83048--83077, 2025{\natexlab{b}}.

\bibitem[Yin et~al.(2025)Yin, Zhang, Zhang, Freeman, Durand, Shechtman, and
  Huang]{yin2025slow}
Tianwei Yin, Qiang Zhang, Richard Zhang, William~T Freeman, Fredo Durand, Eli
  Shechtman, and Xun Huang.
\newblock From slow bidirectional to fast autoregressive video diffusion
  models.
\newblock In \emph{2025 IEEE/CVF Conference on Computer Vision and Pattern
  Recognition (CVPR)}, pages 22963--22974. IEEE, 2025.

\bibitem[Zhang et~al.(2026)Zhang, Wu, Liu, Zhong, Wei, Ye, Liu, and
  Liang]{zhang2026muss}
Haojie Zhang, Di Wu, Bingyan Liu, Linjie Zhong, Yuancheng Wei, Xingsong Ye,
  Nanqing Liu, and Yaling Liang.
\newblock Muss: A large-scale dataset and cinematic narrative benchmark for
  multi-shot subject-to-video generation.
\newblock \emph{arXiv preprint arXiv:2604.23789}, 2026.

\bibitem[Zhang et~al.(2025)Zhang, Jiang, Wang, Fang, Zhi, Yan, Kang, Lu, and
  Pan]{zhang2025storymem}
Kaiwen Zhang, Liming Jiang, Angtian Wang, Jacob~Zhiyuan Fang, Tiancheng Zhi,
  Qing Yan, Hao Kang, Xin Lu, and Xingang Pan.
\newblock Storymem: Multi-shot long video storytelling with memory.
\newblock \emph{arXiv preprint arXiv:2512.19539}, 2025.

\bibitem[Zheng et~al.(2025)Zheng, Huang, Liu, Zou, He, Zhang, Gu, Zhang, He,
  Zheng, et~al.]{zheng2025vbench}
Dian Zheng, Ziqi Huang, Hongbo Liu, Kai Zou, Yinan He, Fan Zhang, Lulu Gu,
  Yuanhan Zhang, Jingwen He, Wei-Shi Zheng, et~al.
\newblock Vbench-2.0: Advancing video generation benchmark suite for intrinsic
  faithfulness.
\newblock \emph{arXiv preprint arXiv:2503.21755}, 2025.

\bibitem[Zhou et~al.(2026)Zhou, Gao, Wang, Wang, Wang, Yuan, Li, Ouyang, Shen,
  Lu, et~al.]{zhou2026worldmembench}
Nan Zhou, Jianxiong Gao, Qiuyu Wang, Wen Wang, Jiahao Wang, Yufeng Yuan,
  Shuaiting Li, Hao Ouyang, Yujun Shen, Weiming Lu, et~al.
\newblock Worldmembench: A comprehensive benchmark for memory of video world
  models.
\newblock \emph{arXiv preprint arXiv:2610.19063}, 2026.

\end{thebibliography}
}

\clearpage
\setcounter{section}{0}
\setcounter{table}{0}
\setcounter{figure}{0}
\renewcommand{\thesection}{\Alph{section}}
\renewcommand{\thetable}{S\arabic{table}}
\renewcommand{\thefigure}{S\arabic{figure}}
\renewcommand{\theHsection}{supp.\Alph{section}}
\renewcommand{\theHtable}{supp.\arabic{table}}
\renewcommand{\theHfigure}{supp.\arabic{figure}}

\captionsetup{font=small}
\setlist{nosep,leftmargin=*}

\makeatletter
\renewcommand\paragraph{\@startsection{paragraph}{4}{\z@}%
  {1.2ex \@plus .4ex \@minus .2ex}{-1em}{\normalfont\normalsize\bfseries}}
\makeatother

\flushbottom
\brokenpenalty=10000
\widowpenalty=10000
\clubpenalty=10000

\maketitlesupplementary
\section{Overview}
\label{supp:sec:overview}

This supplement gives the details and analyses that the main paper omits for lack of space.
Table~\ref{tab:map} maps each part of the main paper to the sections of this supplement that expand it.
Sec.~\ref{sec:uec-scope} is a reading guide to Table~4 of the main paper.
Sec.~\ref{supp:sec:construction} describes how the episodes, prompts, and contracts were built, including the full entity granularity rules.
Sec.~\ref{sec:evaluator} gives the complete configuration of the evaluator and each of its stages, from entity discovery to scoring.
Sec.~\ref{sec:reliability} examines the reliability and sensitivity of the scores and the role of unspecified entities in them.
Sec.~\ref{sec:human} gives the protocol and further results of the human study.
Sec.~\ref{sec:examples} follows one episode from candidate issues to its score and counts the confirmed issues by subcategory.
Sec.~\ref{sec:limitations} discusses limitations and ethics and lists the materials that will be released.

Tables and figures of this supplement are numbered S1, S2, and so on, and numbers without the S prefix refer to the main paper.
Unless marked as coming from a development run, every value comes from the evaluation runs of the main paper, which Sec.~\ref{sec:config} describes.
Table~\ref{tab:scope} lists the videos that each analysis covers.

\begin{table}[h]
\caption{\textbf{Where this supplement expands the main paper.}}
\label{tab:map}
\centering
\footnotesize
\renewcommand{\arraystretch}{1.26}
\begin{tabularx}{\linewidth}{@{}P{0.42\linewidth}L@{}}
\toprule
Main paper & This supplement \\
\midrule
Sec.~3.2, full entity granularity rules & Sec.~\ref{sec:granularity} \\
Sec.~3.2, episodes and contracts & Secs.~\ref{sec:episodes}--\ref{sec:contract-example} \\
Sec.~3.3.1, discovery details & Sec.~\ref{supp:sec:discovery} \\
Sec.~3.3.2, requirement list and answering rules & Sec.~\ref{sec:vp} \\
Sec.~3.3.2, all checklist items & Sec.~\ref{sec:vv}, Table~\ref{tab:checklist} \\
Sec.~3.3.2, unspecified-object check & Sec.~\ref{sec:uoc} \\
Secs.~3.3.3--3.3.4, verification and scoring & Secs.~\ref{supp:sec:verification}--\ref{supp:sec:scoring} \\
Table 1, independent re-examination & Sec.~\ref{supp:sec:verification} \\
Sec.~4.1, evaluator details & Secs.~\ref{sec:config}, \ref{sec:repair}, \ref{sec:cost} \\
Sec.~4, stability across repeated runs & Sec.~\ref{sec:repeats} \\
Table 4, 95\% intervals & Sec.~\ref{sec:ci}, Table~\ref{tab:ci} \\
Sec.~4.2, specified vs.\ unspecified entities & Secs.~\ref{sec:density}--\ref{sec:proxy} \\
Sec.~4.3, human study and baselines & Sec.~\ref{sec:human} \\
Sec.~5, limitations & Sec.~\ref{sec:limitations} \\
\bottomrule
\end{tabularx}
\end{table}

\begin{table}[t]
\caption{\textbf{Videos covered by each analysis.} Five models: Wan2.1, Wan2.2, CineTrans-DiT, HoloCine, StoryMem. Human-study set: 24 episodes of the five models.}
\label{tab:scope}
\centering
\scriptsize
\begin{tabularx}{\linewidth}{@{}LP{0.36\linewidth}@{}}
\toprule
Analysis & Videos \\
\midrule
Table 4 (scores, unspecified shares) & 1,600 (all eight models) \\
Intervals for Table 4 (Sec.~\ref{sec:ci}) & 1,600 (all eight models) \\
Repeated runs (Sec.~\ref{sec:repeats}) & 120 (human-study set) \\
Other evaluation models (Sec.~\ref{sec:repeats}) & 120 (human-study set) \\
Scoring sensitivity (Sec.~\ref{sec:sensitivity}) & 1,000 (five models) and 120 \\
Episode length (Sec.~\ref{sec:length}) & 1,000 (five models) \\
Unspecified-object check ablation (Sec.~\ref{sec:ablation}) & 1,000 (five models) \\
Specified-only proxy (Sec.~\ref{sec:proxy}) & 1,600 (all eight models); 120 (human-study set) \\
Human agreement (Sec.~\ref{sec:human}) & 120 (human-study set) \\
\bottomrule
\end{tabularx}
\end{table}

\subsection{Reading Table~4}
\label{sec:uec-scope}

Table~4 of the main paper reports three dimension scores, their mean (the UEC score), the unspecified shares, and the shares of entities with issues.

\paragraph{Dimension scores.}
Video-prompt alignment checks the requirements of the contract, 165 per episode on average (Sec.~\ref{sec:contract-example}).
Intra-shot and inter-shot consistency cover every entity in the ledger, specified or unspecified.
Issues of both kinds of entity deduct by the same rule (Sec.~\ref{supp:sec:scoring}).

\paragraph{UEC.}
The benchmark is named UEC-Bench to highlight its core contribution, the systematic inclusion of unspecified entities in the evaluation.
Its overall score takes its name from the benchmark: ``UEC score'' denotes the score that UEC-Bench reports, and like any abbreviation used as a name it is not read as its expansion; it is not a score of unspecified-entity consistency alone.
As the main paper's Fig.~1 states, UEC-Bench ``scores prompt alignment, intra-shot consistency, and inter-shot consistency over specified and unspecified entities alike and averages the three into the UEC score''.
The three dimension scores are averaged with equal weights.
Sec.~\ref{sec:sensitivity} shows how the ranking of the models changes when the dimensions are weighted differently.
Video-prompt alignment checks only the requirements of the prompt, while the two consistency dimensions cover specified and unspecified entities alike.
Specified content therefore has the larger influence on the UEC score.
The name of the score does not give unspecified entities more weight.
Separate measures, described next, report the unspecified-entity part of the evaluation.

\paragraph{Unspecified-entity measures.}
The two unspecified shares are the share of unspecified entities among all entities and the share of confirmed intra-video issues that have an unspecified subject.
The columns ``Entities with issues'' give the share of specified and of unspecified entities that are a subject of at least one confirmed intra-video issue.
An evaluation that ``checks only specified entities'' (main paper, Fig.~1) does not measure the shares that concern unspecified entities.
Results with UEC-Bench are reported as in Table~4, with all these columns together.
Table~\ref{tab:density} gives the counts behind these columns per model, and Table~\ref{tab:cross-split} splits the inter-shot issues by subject type.
Table~\ref{tab:ci} gives the 95\% intervals of the scores.

\paragraph{Effect size.}
When the issues of unspecified entities are left out (the specified-only proxy of Sec.~\ref{sec:proxy}), each model's inter-shot score rises by 1.0 to 4.5 points and its UEC score by 0.4 to 2.1 points (Table~\ref{tab:proxy}).
This reverses the UEC order of the two closest pairs of Table~4, HoloCine against Wan2.2 (86.3 and 85.4) and ShotStream against Wan2.1 (78.0 and 77.3).

\section{Benchmark Construction}
\label{supp:sec:construction}

\subsection{Episodes and prompts}
\label{sec:episodes}

\paragraph{Source windows.}
An episode is a window of 4, 6, or 8 consecutive shots of one source video from the SF20K~\cite{ghermi2026long} or MuSS~\cite{zhang2026muss} index.
Each shot is to be generated as about five seconds of video.
A scene-change detector proposes the shot boundaries, which are checked on the original frames when the window is reviewed.
Before windows are drawn, a GPT-6 Astra call screens contact sheets of each source video for promotional cards, advertisements, end credits, and title-only segments.
This screening call is separate from the prompt writing and checking described below.
A contact sheet holds the middle frame of every proposed shot, the opening frames, and frames of the last 90 seconds sampled every 0.5\,s.
Every window stays clear of each flagged interval by a margin of 1\,s on each side.
A source work contributes at most three non-overlapping windows.

\paragraph{Prompt writing and checking.}
For each window, GPT-6 Astra (medium effort) receives at least five frames per shot, taken at the start, one quarter, the middle, three quarters, and the end of the shot, with more frames for long shots.
It records the visible facts with frame references and writes one global prompt and one prompt per shot.
Camera language (shot scale, angle, composition, movement) is added only where the source frames support it, with no fixed proportion of shots.
Camera movement also requires evidence across time.
A second GPT-6 Astra call with a fresh context then checks the prompt against the same frames.
It checks the stated facts, action durations, character identity and presence, and the physical setting.
It also checks whether each window shot is a single shot and whether the source is free of overlays.
A window is kept only when the writing call judged it suitable, the checking call passed it, and every required check was supported.
This procedure produced 400 windows, 200 per source.
Of these, 240, 120, and 40 have 4, 6, and 8 shots, respectively.

\paragraph{The 200 evaluated episodes.}
The benchmark uses 200 of the 400 windows, chosen before any video was generated.
No generated video or evaluator output was used in the choice.
The windows were ranked by a heuristic score within exact quotas, and the first 200 were taken.
The quotas set 100 episodes per source and 120, 60, and 20 episodes with 4, 6, and 8 shots, respectively.
They also set 173 live-action, 24 animated, and 3 documentary episodes and allow at most two episodes per source work.
The 200 episodes come from 152 source works (Table~3).
The heuristic score rewards one setting recurring across shots (up to 50 points), settings whose type naturally holds detail beyond what the prompt lists (20), text cues that make the spatial layout observable (15), recurring source entities (10), and framing not confined to faces or extreme close-ups (5).
It also down-weights windows dominated by on-screen game content.
The ranking thus favors recurring settings with many objects, where unspecified entities matter.

\paragraph{Generation inputs.}
All models receive the same shot texts.
A model that sees one shot at a time cannot resolve a reference to an entity introduced in an earlier shot (``the Bespectacled man'').
A shot text with such a reference therefore ends with a short description copied verbatim from the earlier shot text.
In the 400 windows, 1,084 of the 2,000 shot texts contain such a reference and received this description.
The description comes entirely from the episode's own prompts, and every shot text keeps its original wording.
Every final shot text fits the 512-token limit of the Wan text encoder without truncation.
Each model then receives these texts in its own input format.
Wan2.1 and Wan2.2 generate each shot as a separate clip from its shot text.
CineTrans-DiT receives the shot texts together with the global prompt as its general prompt.
HoloCine receives the global prompt and the shot texts joined with its native shot separators.
Its text length is raised to 1,519 tokens so that every episode fits in full.
StoryMem receives the texts in its story-script format, with every boundary marked as a cut.
ShotStream, MultiShotMaster, and LongLive receive the shot texts with an empty global caption.
The evaluator always reads the original global and shot prompts, not these adapted inputs.

\subsection{Contract format and compilation}
\label{sec:contract-format}

\paragraph{Fields.}
Each contract has three parts (Figure~2).
\emph{Entity descriptions} give every entity an integer ID, a type (character, location, or object), a name, and one description that gathers every stable attribute the prompt states anywhere in the episode.
\emph{Shot blocks} describe each shot.
A block gives the location of the shot, which is only its main setting.
Two locations are given only when the camera moves from one to the other within the shot.
The block also gives the camera view, camera movement, and visual style of the shot, each empty when the prompt states none.
It lists the entities expected to be visible (\emph{presence}, locations included) and the entities the prompt says are not visible (\emph{absent}).
Three lists of requirements follow: \emph{appearance}, \emph{spatial}, and \emph{motion}.
The appearance list holds only what changes or is emphasized in this shot.
Apart from these entries, the appearance of an entity is its description.
Every boundary between consecutive shots is a required cut (Table~3).
\emph{Cross-shot requirements}, the third part, compare two or more named shots, including how the camera changes between them (``Shot4 is a reverse view of Shot3'').
Every requirement names its entities and an expected state.
It has a quantifier, which is \code{exists} (at least one frame), \code{throughout} (every frame showing the entity), \code{start} or \code{end} (the first or last frames of the shot), \code{when\_visible} (every frame showing the entity must agree), or \code{cross\_shot}.
Spatial and motion requirements also carry a predicate and a coordinate frame, which is \code{entity} (the entity's own left or right), \code{screen}, \code{camera}, \code{world}, or \code{none}.

\paragraph{Rules.}
A contract is treated as ground truth, so it contains only requirements that the prompt fixes unambiguously.
Wording with more than one reasonable reading produces no requirement.
Identity is the one exception: whether mentions in different shots denote the same entity is decided by the most likely reading in context, because inter-shot consistency is checked against this decision.
A detail the prompt does not mention is not forbidden.
The remaining rules concern directions, camera information, lighting, and style.
When the subject is a person and no frame is stated, left or right is the person's own side.
In contrast, positions within the frame composition, facing direction, and entering or leaving the view stay in screen terms.
Focus, softness, or blur of part of the image is camera information of the shot whose text states it (camera view).
An instruction to hold the shot or view is a static camera (camera movement).
Lighting that is stated in every shot of a place, stated globally, or explicitly carried over is part of the description of the place.
Any other stated lighting is a requirement of its shot.
A style is recorded only when the prompt states a visual or rendering style in its own words.
A global style applies to every shot unless a shot states its own.

\paragraph{Compilation and review.}
Claude Sonnet agents compile the contracts from the prompt text alone, without source frames or generated videos.
They follow written rules and start from an earlier automatically compiled claim list.
The agents issue edit operations rather than writing contracts directly.
A program applies the operations and accepts the result only when every entity, requirement, and camera value is backed by a literal quote of the prompt.
Every reference must also resolve, and the presence lists must be \mbox{consistent}.
A fresh agent then reviews every episode with an identity decision or a change.
It sees the original contract and the decisions, but not the reasons of the compiling agent.
A separate adjudication pass under the same rules resolves disagreements.
Each contract has two versions, with and without the supporting prompt quotes of every field.
The evaluator receives only the latter, model-facing version.

When the contracts were first converted to the current format, a fresh agent checked the contracts of all 400 windows against their prompts and the rules.
It revised 247 contracts and recorded 529 problems, and every later pass therefore includes an independent review.
In the later cleanup of the 200 benchmark contracts, the agents made 628 identity decisions (262 same entity, 366 separate entities).
The cleanup review checked the identity decisions and changes of the 197 episodes that had any, and it disagreed on 22 items.
Of these, 19 were resolved as the review proposed and 1 as the compiling agent had decided.
The other 2 concerned a person's left or right and took the person's own side, as the rules above state.
In the same cleanup, a program resolved 811 of the 2,124 left/right mentions, and agents resolved the rest.
The program covered three cases: no direction left in the requirement, a body side, or a frame stated in the prompt.
The cleanup also rewrote 112 requirements into the person's own frame.
The granularity rules of Sec.~\ref{sec:granularity} were applied in a separate pass with the same structure (compilation, independent review, and adjudication).
This pass added 1,289 worn or carried items and 214 place parts as entities, which gives the 4,438 entities of Table~3.
The authors fixed the boundaries of the granularity rules through rulings on 26 example cases.
They also checked sample contracts at each revision of the contract format and compilation rules.

\subsection{Entity granularity}
\label{sec:granularity}

The same rules decide what is an entity in a contract and in a generated video (Table~\ref{tab:granularity}), except that a contract holds only the entities the prompt names, while discovery also finds unnamed ones.
An entity that a contract holds is \emph{specified}.
In the generated video, an entity is specified when it is bound to a contract entity (Sec.~\ref{supp:sec:discovery}) and unspecified otherwise.
Entities that the detection agents or the unspecified-object check newly create are unspecified.
An object that the check assigns to an existing ledger entity takes the label of that entity (Sec.~\ref{sec:uoc}).

\begin{table*}[t]
\caption{\textbf{Entity granularity in contracts and in entity discovery.} A contract lists an entity only when the prompt names it, and lists it in a shot's presence only when that shot's text mentions it. A global statement counts for every shot of the wearer or place. Discovery applies the same rules to every entity visible in the video.}
\label{tab:granularity}
\centering
\footnotesize
\begin{tabularx}{\textwidth}{@{}P{0.17\textwidth}LL@{}}
\toprule
Kind & Contract (prompt-named entities) & Entity discovery (generated video) \\
\midrule
Persons and animals & One character entity each; hair, beard, makeup, and other body features stay in the description. & One entity each; face, hair, hands, and makeup are part of the person. \\
Places & One location entity; a shot lists only its main setting (two when the camera moves between them). A place seen from the setting (a hallway through a door) is in presence, not a shot location. & The room or place as a whole is not a detected entity; its contract location record represents it. \\
Worn or carried items & Each garment, trousers or skirt, footwear, headwear or hair accessory, eyewear, tie or scarf, jewelry or watch, belt, lanyard, badge or strap, glove or apron, and bag is an object entity. A matching pair, a set the prompt names as one item (``a gray suit''), and items a group wears are one entity each. An item named only as missing gets no entity. & Each worn or carried item is a separate entity named by what it is and whose it is (``woman's red scarf''); a matching pair is one entity. Never merged into the wearer. \\
Handled items & An item the prompt describes as handled (raised, moved, used, put down) is an entity. & An item held, handled, or no longer on the person (a cap on a table) is a separate entity. \\
Place parts & Walls, floors, ceilings, ground surfaces (street, snow, grass, sand), floor coverings, trees, bushes, and plants are object entities. A place kept as its own location stays a location; ``tiled'' without saying wall or floor creates none. & Separate entities named by what and where they are (``kitchen back wall''); never merged into the place. \\
Structural parts and fixtures & Doors, doorways, windows, frames, grilles, curtains, pipes, and wall or ceiling fixtures are separate entities, whether or not anything happens to them. & Separate entities. \\
Furniture and props & Separate entities, including background furnishings and small items on a surface. & Separate entities; a seat or back cushion is part of its chair or sofa, and a blanket draped over furniture as a furnishing is part of it; loose throw pillows stay separate. \\
Groups and collections & A group seen as a whole (loose papers, a crowd of guests) is one entity; members the prompt describes separately are separate. & Clearly distinguishable objects stay separate (two cups are two entities); a large number of same-kind items gathered together (a basket of oranges) is one entity even when each item is visible; the container is separate, and separate groups stay separate. \\
Not entities & Sky, clouds, weather, distant landforms. & Sky, clouds, reflections, shadows, light spots. \\
\bottomrule
\end{tabularx}
\end{table*}

\subsection{Example contract}
\label{sec:contract-example}

Figure~2 shows part of the contract for the four-shot episode \#101, which also appears in Figure~4.
The prompt of this episode names 17 entities.
Of these, 7 are worn items or place parts: the Bespectacled man's glasses, jacket, shirt, and knit layer, the Dark-shirted man's necklace and shirt, and the tiled wall.
The glasses of the Dark-shirted man, which StoryMem renders in Shot~2 and drops in Shot~4 (Figure~4), are not in the prompt and therefore not in the contract.
In the generated video, they are an unspecified entity.
The requirement list compiled from this contract (Sec.~\ref{sec:vp}) has 132 entries: 38 presence and 38 description items, 21 spatial, 18 motion, and 8 appearance requirements, 3 camera views, 2 camera movements, 3 cuts, and 1 cross-shot requirement.
Over the 200 episodes, the requirement lists have 32,963 entries, 165 per episode on average.

\section{Evaluator Details}
\label{sec:evaluator}

\subsection{Default configuration}
\label{sec:config}

\paragraph{Models.}
Every vision-language model call of the evaluator, including those of entity discovery, uses GPT-6 Luna (API model \code{gpt-6-luna}) at low reasoning effort through the OpenAI Responses API.
The evaluator also uses two detectors.
WeDetect-Anything~\cite{fu2025wedetect} runs with its official inference defaults: objectness 0.1, IoU 0.7, at most 100 proposals per frame, and a class-match score of at least 0.4.
Grounding DINO~\cite{liu2024grounding} (Swin-T) uses box and text thresholds of 0.30 and 0.25.
DINO appearance features are computed for each entity occurrence, but no stage or score~reads~them.

\paragraph{Runs.}
The run of Table~4 of the main paper consists of two evaluations with this configuration.
One is the main run (run 0) on the 1,000 videos of the first five models (Wan2.1, Wan2.2, CineTrans-DiT, HoloCine, and StoryMem), and the other is the evaluation of the 600 videos of the three added models (ShotStream, MultiShotMaster, and LongLive).
The three models were evaluated after the main run, so analyses made on the main run alone cover the first five models (Table~\ref{tab:scope}).
Runs 1 and 2 repeat the evaluation on the 120 human-study videos (Sec.~\ref{sec:repeats}).

\paragraph{Keyframes.}
Each shot has six keyframes.
When a model generates one clip per shot, the keyframes are uniformly spaced from the clip's start to $\max(0.25\,\mathrm{s}, 0.05\,T)$ before its end, where $T$ is the clip duration.
An episode video is cut into shots at the cut or prompt-switch frames recorded at generation time, rather than on a nominal five-second grid, and a missing or inconsistent record is an error.
In an episode video, keyframe sampling starts at frame 0 in the first shot.
In each later shot, it starts four frames after the recorded cut.
If a clearly visible cut lands at or after that frame and within the first eight frames of the shot, sampling starts on the frame after the visible cut instead.
The offset is needed because the first four frames of a shot can still show the previous shot.
The last keyframe lies four frames before the shot ends.

\paragraph{What each call sees.}
Each attached image is preceded by a text label such as ``Shot3 frame 2''; pixels and order are unchanged.
Shot clustering, prompt binding, background discovery, and the inventory step of the unspecified-object check (Sec.~\ref{sec:uoc}) see only the six keyframes of their own shot.
Cross-shot matching sees all keyframes, the original prompt, and each shot's location.
Both detection agents and both verifiers see all keyframes, the original prompt, and the model-facing contract.
Severity calls see only text.

\begin{table}[t]
\caption{\textbf{Model calls per episode} with $N$ shots and $K$ suspects (objects verified by the unspecified-object check, Sec.~\ref{sec:uoc}), without repair rounds. A stage skips its call when it has no candidate or issue to judge. Sec.~\ref{sec:cost} gives the measured mean.}
\label{tab:calls}
\centering
\footnotesize
\begin{tabularx}{\linewidth}{@{}Lr@{}}
\toprule
Stage & Calls \\
\midrule
Shot clustering (two passes) & $2N$ \\
Prompt binding (two passes) & $2N$ \\
Background discovery and box choice & $2N$ \\
Cross-shot matching & 1 \\
Video-prompt detection & 1 \\
Intra-video detection (one per checklist item) & 19 \\
Unspecified-object check: inventory / identity / per suspect & $N$ / 1 / $K$ \\
Video-prompt verification & 1 \\
Intra-video verification & 2 \\
Severity scoring (one per kind of issue) & 2 \\
\bottomrule
\end{tabularx}
\end{table}

\subsection{Entity discovery and prompt grounding}
\label{supp:sec:discovery}

\paragraph{Detection and clustering.}
WeDetect-Anything proposes boxes on every keyframe.
For each shot, a clustering agent receives the six keyframes and the candidate list, with the ID, label, box, and score per frame.
It assigns every candidate exactly once, either to a named entity or to a rejection with a reason.
The agent follows the discovery column of Table~\ref{tab:granularity}, forms entities only from detector candidates, and keeps every box unchanged.
In each keyframe, an entity takes the highest-scoring box among its candidates.

\paragraph{Prompt binding and recovery.}
A binding agent then sees the shot's keyframes, its entities, and the contract entities, and marks each local entity as bound to one contract entity or as unbound.
Binding means identity: a local entity with a wrong color or changed text can still bind, whereas an ambiguous candidate or a part-whole mismatch stays unbound.
Several local entities may bind to one contract entity whose name or description states several items or members (``Three glasses'').
The agent also writes a short category query for each contract entity, which the recovery step uses.
In every keyframe where a contract entity of the shot's presence list has no box, Grounding DINO searches with that query and attaches the highest-scoring box inside the image.
A keyframe that still has no box counts as unobserved.

\paragraph{Background objects.}
The detector often proposes no box at all for small and flat background objects.
After the first pass of clustering, binding, and recovery, one agent per shot therefore lists the clearly visible objects that no entity of the first pass covers.
Each object comes with a detector query and the keyframes that show it.
Grounding DINO searches for each object on those keyframes.
In each keyframe, a second agent then picks the drawn box that covers the object, or none (for example, when a box covers the whole frame or another object).
The chosen boxes join the original detections, and clustering, binding, and recovery run again on the combined set.

\paragraph{Cross-shot matching.}
Specified entities are linked across shots by their contract IDs.
The unbound occurrences go to one matching agent.
The agent sees all keyframes together with the prompt and the location of each shot.
In the keyframes, specified occurrences are drawn as labeled blue boxes and unbound ones as labeled orange boxes.
Each unbound occurrence gets exactly one of three destinations.
It is merged with unbound occurrences of other shots into one unspecified entity, kept alone as an unspecified entity, or merged into a specified entity whose binding was missed.
Identity follows the most likely reading of the episode, which weighs appearance together with the setting, the position in the place, the relations to other entities, the continuity of actions, and the story; no single cue decides.
An appearance difference alone does not split an entity, since later stages judge such differences.
Occurrences are kept apart only on positive evidence.
Positive evidence here means that both are visible in one shot, the prompt introduces another one, they are in different places with no movement between them, or they have different roles in the story.
An unspecified entity has at most one occurrence per shot.
Worn items and place parts always stay separate from their wearer or place.

\paragraph{Ledger.}
The output of entity discovery is the entity ledger, with one list of specified entities and one list of unspecified entities.
Each record has an ID, the contract ID of a specified entity, a name, a short description, and the entity's occurrences (shot, keyframes, and boxes).

\begin{table*}[t]
\caption{\textbf{Issue subcategories} of the three dimensions. Subcategories marked $^{\circ}$ may have no subject entity.}
\label{tab:subcats}
\centering
\scriptsize
\begin{tabularx}{\textwidth}{@{}P{0.16\textwidth}L@{}}
\toprule
\multicolumn{2}{@{}l}{\textbf{Video-prompt alignment}} \\
presence\_count & a required entity or setting is missing; wrong number of instances; an entity required to stay offscreen is visible; the shot is black or unreadable \\
attribute & a described identity descriptor or property (age, gender, hair, clothing, accessories, colors, materials, legible text, object state) is wrong, including one shown on the wrong character \\
action\_state & a described action, motion, pose, gesture, or expression is missing, wrong, incomplete, reversed, out of order, or has the wrong start or end state \\
interaction & the wrong actor or object, or a described holding, touching, using, giving, or contact relation is absent or wrong \\
spatial\_orientation & described positions (left/right, foreground, between, on/in/under), distances, facing, or gaze directions \\
setting\_environment$^{\circ}$ & described location type, background elements, time of day, weather, season, lighting, or mood \\
camera\_composition$^{\circ}$ & described shot scale, camera angle, viewpoint, framing, focus, or camera movement \\
style\_rendering$^{\circ}$ & described rendering style, animation or live action, color mode, aspect \\
stated\_continuity & an explicit sameness phrase (same, remains, returns to, still, again) is contradicted and no more specific subcategory fits \\
shot\_structure$^{\circ}$ & a required cut or shot plan is not realized (no cut, a shot's content in another shot's interval, a new shot repeating the previous one) \\
\midrule
\multicolumn{2}{@{}l}{\textbf{Intra-shot consistency}} \\
identity\_drift & a person or animal morphs into a different-looking individual \\
appearance\_drift & clothing, hair, accessories, colors, or marks change with no visible action \\
action\_discontinuity & pose, position, or action progress jumps, reverses, or restarts between frames \\
object\_instability & an object's type, shape, size, color, count, text, or state changes with no action \\
presence\_instability & an entity pops in or out without entry, exit, or occlusion; an instance duplicates, splits, or fuses with another \\
background\_drift & background elements or layout change in a way camera motion does not explain \\
physical\_integrity & malformed or fused bodies, hands, or faces; extra or missing limbs; interpenetration; a reflection or shadow without source \\
relation\_break & a held object teleports or switches hands; contact, support, or relative position changes with no action \\
unintended\_cut$^{\circ}$ & the camera setup, composition, or setting jumps discontinuously inside one shot \\
rendering\_instability$^{\circ}$ & flicker, exposure flash, black or blank frames, or a style switch inside the shot \\
other$^{\circ}$ & any other visible inconsistency, only from the open-ended sweep (O1), described in the agent's words \\
\midrule
\multicolumn{2}{@{}l}{\textbf{Inter-shot consistency}} \\
character\_identity & the same character's face, age, gender, skin tone, or build differs \\
character\_appearance & the same character's hair, facial hair, clothing, accessories, makeup, or injuries differ, including attributes that migrate to another character \\
object\_identity & the same prop, object, vehicle, or animal differs in type, shape, color, markings, text, or size \\
setting\_background & a recurring location's architecture, wall and floor materials, doors, windows, furniture, vegetation, signage, or layout differ \\
environment\_lighting$^{\circ}$ & the same continuous scene changes time of day, weather, season, lighting, or grade without a supported cause \\
rendering\_style$^{\circ}$ & the visual style differs across shots that should share one style \\
instance\_count & a recurring entity is duplicated, merged, or replaced; two distinct characters look identical or swap looks \\
state\_action\_continuity & pose, position, held objects, object or clothing state, or action progress is incompatible across a boundary or a return \\
spatial\_continuity & relative positions, left/right arrangement, or screen direction flip without a cause \\
other$^{\circ}$ & any other visible inconsistency, only from the open-ended sweep (O1) \\
\bottomrule
\end{tabularx}
\end{table*}

\begin{table*}[t]
\caption{\textbf{Checklist of the intra-video alignment agent} (one call per item, wording condensed from the prompt). Each item sets a minimum: the agent compares every entity, shot, and frame that the item covers.}
\label{tab:checklist}
\centering
\scriptsize
\begin{tabularx}{\textwidth}{@{}P{0.03\textwidth}P{0.14\textwidth}LP{0.16\textwidth}@{}}
\toprule
Item & Title & What is compared & Subcategories \\
\midrule
W1 & Camera position and movement & For each keyframe the shot scale, angle, and camera position; each change is none, a smooth movement (the background slides continuously and stays aligned), a gradual reframe, or a jump. A jump inside a shot is an unintended cut unless the text asks for it; large subject motion against an aligned background is motion. & unintended\_cut, background\_drift \\
W2 & Clothing, hair, accessories & For every visible character, each garment (type, color, pattern, logo, neckline, sleeves, layers), hairstyle and color, headwear, glasses, jewelry, carried items, frame by frame; a change needs a shown action. & appearance\_drift \\
W3 & Face and identity & Face shape, apparent age, skin tone, eyes, nose, mouth, facial hair, makeup, marks; a different-looking person in the same role is identity drift even with the same clothing; expression, head turn, and lighting are normal. & identity\_drift \\
W4 & Actions, poses, bodies & Pose and position evolve continuously (no teleport, instant pose swap, reversal, or restart); hands, limbs, and bodies stay intact. & action\_discontinuity, physical\_integrity \\
W5 & Objects, held items, contacts & Each object keeps type, shape, size, color, count, text, and state; held objects do not teleport or switch hands; contact and support do not change without an action. & object\_instability, relation\_break \\
W6 & Presence and count & Nobody and nothing pops in or out without entry, exit, or occlusion; no instance duplicates, splits, or fuses. & presence\_instability \\
W7 & Background, lighting, rendering & Background structures and layout stay the same apart from camera motion; no flicker, exposure flash (check the first keyframe), black or blank frame, or style switch. & background\_drift, rendering\_instability \\
W8 & Background objects & List the small and background objects of the shot, whether or not the ledger or the text has them (items on walls, shelves, tables, counters, windowsills; signs, switches, lamps; furniture and plants; things held); none may appear, disappear, move, or change shape, color, size, or count without a visible cause. & object\_instability, presence\_instability \\
\midrule
X1 & Camera positions across shots & A new camera position at a cut is normal. Shots showing the same place from a comparable position, including returns, must match in background and layout; across positions the visible layout must be explainable by the camera change. & setting\_background, spatial\_continuity \\
X2 & Faces of recurring characters & Every character seen in two or more shots, including secondary, back-view, and over-the-shoulder figures: face shape, age, skin tone, ethnicity, build, features, across all its shots. & character\_identity \\
X3 & Clothing, hair, accessories & Each recurring character's garments, hairstyle, facial hair, headwear, glasses, jewelry, makeup across all its shots; no attribute migrates to another character. & character\_appearance \\
X4 & Actions across boundaries & At each boundary and each return, the last keyframe of the earlier shot against the first of the next: pose, position, held objects, clothing state, action progress; an action must not reset unless the shots are separated in time. & state\_action\_continuity \\
X5 & Recurring objects & Type, shape, color, markings, text, and size of every recurring prop, vehicle, animal, or furniture item. & object\_identity \\
X6 & Recurring settings & Architecture, wall and floor colors and materials, doors, windows, furniture, vegetation, signage, set dressing; each changed element is named as an entity; a new camera direction explains new elements only if they match the place's style and materials. & setting\_background \\
X7 & Lighting and style & Lighting, time of day, weather, season, grade, and rendering style across shots of one scene, and across the episode when it should share one style. & environment\_lighting, rendering\_style \\
X8 & Instances & Recurring characters and objects stay distinct and singular: no duplicate, merge, replacement, or look swap. & instance\_count \\
X9 & Layout and screen direction & Relative positions, left/right arrangement, and screen direction of recurring entities stay consistent unless an action or the camera change explains them. & spatial\_continuity \\
X10 & Background objects of a shared view & For shots showing the same part of a place from a comparable view, the background objects of that view (furniture, wall items, signs, switches, plants, shelves and their contents) must be visible at the same place unless the camera change, occlusion, or an action explains it, and keep type, color, shape, and size. & object\_identity, setting\_background \\
\midrule
O1 & Open-ended sweep & Every frame is re-examined for any other obvious or meaningful consistency problem; a listed subcategory is used if it fits, otherwise \code{other} with a free description. & any \\
\bottomrule
\end{tabularx}
\end{table*}

\subsection{Video-prompt alignment}
\label{sec:vp}

\paragraph{Requirement list.}
The requirement list is compiled deterministically from the contract, once per episode, with one entry and a unique ID per item.
For each shot, every entity in its presence list has a presence item (\code{S$n$-pre-$e$}, for shot $n$ and entity $e$) and a description item (\code{S$n$-des-$e$}).
The description item checks the episode-wide description of the entity in this shot.
The shot's requirements about that entity follow these two items.
An entity in the shot's absent list gets an absence item instead.
Next come the shot's requirements relating several entities, then its camera view, camera movement, and style when set, and finally the cut from the previous shot (\code{S$n$-cut}).
Each cross-shot requirement is attached to the last shot it names.

\paragraph{Inputs.}
The video-prompt detection agent makes one call per episode (Table~\ref{tab:calls}).
It receives all keyframes, the requirement list, the model-facing contract, the contract entities with their ledger IDs, and the ledger as reference.
It is also told whether each shot is a separate clip or an interval of one episode video.
The original prompt is context only.
The agent checks only what the contract requires.

\paragraph{Answer format.}
The output schema is built per call with one required answer per requirement ID, in list order, so that no requirement can be skipped, repeated, or invented.
Each answer has three parts, written in this order.
(1) \emph{Observed}: what the deciding frames show, with concrete values (each described attribute as rendered, where entities are relative to each other, what happens from the first to the last keyframe, the shot scale and camera movement).
An observation that restates the requirement is invalid.
(2) \emph{Verdict}: satisfied, violated, not verifiable (the deciding content is invisible, too small, occluded, or ambiguous in every frame), or, for a description only, not applicable (the shot's own requirements describe the entity differently).
(3) For a violation, its \emph{problems}: what the frames show instead, the frames showing it, a subcategory (Table~\ref{tab:subcats}), and a confidence.
A description or camera item can fail in several independent attributes.
Each failed attribute is then a separate problem that quotes the contradicted words literally.
The same failed attribute in several shots uses the same words.

\paragraph{Judging rules.}
Quantifiers are judged as defined in Sec.~\ref{sec:contract-format}, and directions in the requirement's coordinate frame.
A description applies in every shot where the entity appears, unless that shot's own requirements describe it differently, that is, give another value of the same attribute or state a change.
A shorter mention keeps the description in effect.
A shot that satisfies the words violates nothing, even if it looks different from another shot, while explicit continuity words (``the same sofa'', ``remains'', ``returns to'') make sameness itself the requirement.
A camera, crop, or lighting difference counts as a violation only when the contract constrains it.
A cross-shot requirement is judged by comparing the named shots, with frames from each.
A cut must show a clear shot change.
In an episode video, a cut that lands a fraction of a second late, so that only the first keyframe still shows the previous shot, is satisfied.
An absence item is satisfied when no frame shows the entity and violated when any frame does.
When a required entity is missing, every other requirement of the shot that needs the entity to be seen (appearance, position, relation, action) is also \mbox{violated}.
Each of these violations is reported separately, so that each unmet requirement deducts.
A requirement that holds without the entity is satisfied, and the entity's description is not verifiable.\looseness=-1

\paragraph{Confidence and re-examination.}
A low confidence is allowed only for a real but incompletely resolved contradiction.
A speculative failure is answered satisfied or not verifiable.
An entity may be boxed by the ledger in two or more keyframes of a shot while all requirements about it in that shot are answered not verifiable.
In that case, the answers are returned to the agent once for re-examination.

\paragraph{From answers to candidates.}
Each problem becomes one candidate issue that states the requirement words and what the frames show.
Candidates that report the same failure of the same requirement are merged by rule.
The rule requires the same requirement, the same quoted words, and the same subcategory, and it counts an entity description as one requirement across shots.
A not-verifiable answer creates no candidate and deducts nothing.
In the 120 human-study videos, 906 of 19,855 answers (4.6\%) were not verifiable, 1,959 violated, and 16,990 satisfied.

\subsection{Intra-video alignment}
\label{sec:vv}

\paragraph{Calls.}
Intra-video detection makes one call per checklist item (Table~\ref{tab:checklist}), 19 per episode: eight intra-shot items (W1--W8), ten inter-shot items (X1--X10), and the open-ended sweep (O1).
Each call receives all keyframes, the full ledger with occurrences and boxes, and whether each shot is a separate clip or an interval of one episode video.
It also receives the original prompt and the contract, which serve only to tell which entities should be the same and which changes are caused.
The calls of an episode share instructions, \mbox{inputs}, and keyframes and name their item in a final message, so the shared part can be cached.
O1 also receives the other items, so that it looks for problems outside them.
W1 runs first, and then up to eight calls run in parallel.

\paragraph{What is an issue.}
For its item, the agent finds the entities that the item concerns in the ledger and the keyframes, compares them as the item specifies, and reports every change that no normal cause explains.
The prompt decides whether a change is permissible.
A change that the prompt calls for or allows is normal.
A change of the same entity, place, or moment that the prompt and the frames leave unexplained is an issue, even when the prompt does not describe the changed attribute.
The normal causes fall into three groups.
The first group concerns the camera and the cuts: a smooth camera movement inside a shot, and a different camera position that shows another side of the same entity or place while the elements visible from both still match.
It also includes the shot change itself at a required cut, and a cut in an episode video that lands slightly late, so that only the first keyframe shows the previous shot.
The second group concerns actions, light, and time: an action that the text describes (quoted literally), an action that the frames show, a visible or described light source or time change, and enough elapsed story time shown by the text or clear cues (never for a change of face).
The remaining causes are occlusion, exit, entry, or cropping; observations of different entities; and blur, focus, compression, or small size.
An issue is reported only when the frames show the difference.

\paragraph{Issue format.}
One issue is one physical discrepancy: the same entity changing in the same way between the same shots.
A failure of one entity that recurs across several shots is therefore one issue covering them.
Each issue has a scope, a subcategory of that scope, a description, and its shots.
The scope is intra-shot for W items, inter-shot for X items, and either for O1.
An intra-shot issue has one shot, and an inter-shot issue has two or more.
Each issue also lists its entities, each with a role: \emph{subject} (whose identity, appearance, state, presence, or position is inconsistent) or \emph{context} (which only helps to locate the issue).
It gives key evidence frames for each side of the discrepancy (from at least two shots for an inter-shot issue) and a confidence.
Entities are referenced by ledger ID.
The ID of a contract entity is preferred, and otherwise the ID of an unspecified ledger entity is used.
An entity that the ledger lacks is added with a new key, a name, a description, and the frames that show it.
A worn item or a place part is the subject of a change to it, and its wearer or place is context.
The runner, the program that drives the calls, renumbers the issues and added entities of the 19 calls.
It then checks that the scope follows the item and that shots, frames, and entity keys exist.
It also checks that a subject is present unless the subcategory allows none, that an inter-shot issue has evidence from two shots, and that \code{other} appears only under O1.
Repeated reports of one problem from different items are left to the verifier.

\subsection{Unspecified-object check}
\label{sec:uoc}

The checklist compares many entities per call, and small unnamed objects are easily overlooked.
The unspecified-object check therefore splits the task into steps that each look at one thing.
It makes an inventory of each shot, runs an identity step for the episode, selects suspects by a rule, and verifies each suspect in one call.

\paragraph{Inventory.}
The inventory makes one call per shot.
The call receives the six keyframes and text of the shot, the contract entities, and the specified and unspecified ledger entities of the shot.
It keeps every unspecified ledger entity of the shot and adds every other clearly visible object that the prompt and the contract leave unnamed.
Such objects include small and background things, worn items, and place parts.
Each entry has a name, a description, its location in the frame, the keyframes in which it is visible, and any change within the shot (``none'' or one sentence).

\paragraph{Identity.}
The identity step makes one call per episode with all keyframes and the ledger boxes drawn.
Each listed object outside the ledger is assigned to one of three destinations.
It goes to a ledger entity only when that entity is the same object (a box that contains the object proves nothing).
If no ledger entity is the same object, it joins a new unspecified entity shared with the same object in other shots, or it forms an entity of its own.
Two objects of one shot never share an unspecified entity.

\paragraph{Suspects.}
Suspects then come from a rule, without a model call.
A within-shot suspect is an object whose noted change is not ``none'' or whose visible keyframes have a gap.
Every visible object is also a cross-shot suspect, compared with every other shot that the contract places in the same location.
For each such shot, the rule lists the inventory entries that may be the same object, or notes that nothing similar is listed.
These are entries with the same ledger ID or a shared content word in the name.

\paragraph{Suspect verification.}
Each suspect is verified in one call with every keyframe of the shots involved and the shot texts.
The agent first names an anchor: a fixed thing next to the object's place, visible in both compared frames (a door, a window, a picture, a piece of furniture).
A missing object counts as evidence only when an anchor is named.
The verdict is real, not real, or unclear.
A real verdict comes with a subcategory from four intra-shot and four inter-shot object subcategories, a description, and at least two evidence frames.
Unclear suspects are dropped.
Real findings join the intra-video candidates under item O1, and the intra-video verifier judges them together with the checklist candidates.
The subjects of a finding are the suspect object and the inventory entries that the rule paired with it.
Each subject carries its ledger ID from discovery or from the identity call.
A finding about a specified entity that the inventory listed again therefore counts as specified.

\paragraph{Counts.}
On the 1,000 videos of the first five models, the inventory kept 9,491 entries of ledger entities and listed 3,889 new objects.
The identity call assigned 257 of these new objects to specified entities, 416 to unspecified ledger entities, and the remaining 3,216 to 2,974 new entities.
The rule produced 14,428 suspects (1,734 within a shot, 12,694 across shots).
Because the rule makes every visible object a suspect against every other shot of its location, most suspects are expected to be consistent.
The suspect verification is the filter, and only suspects judged real reach the intra-video verifier.
The suspect verification judged 2,149 suspects real, 6,687 not real, and 5,592 unclear.
Of the resulting issues, the intra-video verifier confirmed 777, rejected 895, and left 207 uncertain.%
\footnote{These counts are from the run before a correction of the suspect rule. In 97 videos, the rule had skipped 49 of the 12,743 cross-shot suspects, and the correction changed the inputs of 230 other suspects. The corrected run verified these 279 suspects again. In 28 of the 97 videos, the new verdicts changed the candidates, so the intra-video verification and scoring were rerun, and the scores of 26 videos changed. Table~4 and every score use the corrected run; model means moved by at most 0.08 UEC points.}

\subsection{Verification}
\label{supp:sec:verification}

\paragraph{Video-prompt alignment.}
Each candidate states one failed requirement, and repeated reports were already merged by rule.
The verifier therefore makes a single verdict call.
It receives the issues, each with its subcategory, description, shots, quoted requirement, frames, entities, and the requirement items concerned.
It also receives the model-facing contract as the source of every requirement, the original prompt as context, all keyframes, and only the ledger records of the entities that the issues name.

\paragraph{Intra-video alignment.}
The intra-video verifier makes two calls.
The first call keeps, splits, or merges the candidates and rewrites them, and it leaves the verdicts to the second call.
A candidate that bundles independent discrepancies (different entities, attributes, or shot pairs) is split.
Candidates that report the same concrete discrepancy of the same entity over the same span in the same dimension are merged, so that the problem deducts once.
An intra-shot candidate is never merged with an inter-shot candidate.
Each rewritten issue states one discrepancy clearly and corrects unclear or partly wrong details without weakening the core problem, the discrepancy it reports.
It also records the candidates it comes from, and the runner names it after them: a kept candidate keeps its name (V3), merged candidates are joined (V1+V4), and the parts of a split candidate are numbered (V2.1).
The second call gives the verdicts and sees only the rewritten issues, not the candidates.

\paragraph{Verdicts.}
Both verifiers judge the core problem of each issue.
Inaccurate secondary details are never a reason to reject an issue.
Each verdict has a reason from the fixed list of Table~\ref{tab:basis}, evidence frames with the role supports, refutes, or context, a confidence (high, medium, or low), and a short assessment.
The schema of every verdict call has one required entry per issue ID, so that no issue can be skipped, repeated, or invented.
On the 120 human-study videos (run 2 of Sec.~\ref{sec:repeats}), the video-prompt verifier confirmed 1,511 of 1,856 candidates, rejected 195, and left 150 uncertain.
In the same run, the intra-video verifier rewrote 782 candidates into 681 issues, of which it confirmed 368, rejected 272, and left 41 uncertain.

\paragraph{What independent re-examination means.}
Table~1 of the main paper lists independent re-examination of every finding as a property of UEC-Bench.
The property means that every candidate, including the findings of the unspecified-object check, is judged in a separate verification call that shares no context with the detection call.
The verdict call receives the issue statement, the cited frames, the entities, the contract, and the keyframes.
It does not receive the detection agent's notes, its answers to other requirements or items, or its confidence.
For intra-video issues, it sees only the rewritten issue and not the original candidate.
It must choose a reason from the fixed list of Table~\ref{tab:basis} and cite frames that support or refute the core problem.
Its output must also pass the program checks.
For example, a confirmed verdict without a supporting frame in the issue's shots is returned for repair.
Detection and verification use the same model at the same effort (Sec.~\ref{sec:config}).
The effect of the re-examination is visible in the verdicts.
In run 2, 11\% of the video-prompt candidates and 40\% of the intra-video issues were rejected.
Two of the four candidates of Table~\ref{tab:trace} were also rejected.

\subsection{Scoring}
\label{supp:sec:scoring}

Two severity calls, one per kind of issue, receive only the confirmed and uncertain issues, as text.
Each issue is sent with the verifier's statement, shots, quoted requirement or comparison, verdict, reason, confidence, assessment, entity names and roles, and evidence observations.
Rejected issues are not sent and deduct nothing.
The severity calls assign each issue one of four severities (Table~\ref{tab:severity}).
An uncertain issue is rated as if it were real, since its uncertainty enters through a separate factor.
Severity is likewise rated apart from the verifier's confidence, which has its own factor.
An issue is judged at the full extent the verifier describes, so a split issue is neither shrunk nor inflated by the other parts of its candidate.
An inconsistency is rated the same whether or not its entity appears in the text.

\begin{table}[t]
\caption{\textbf{Fixed reasons of the verifiers.}}
\label{tab:basis}
\centering
\scriptsize
\begin{tabularx}{\linewidth}{@{}P{0.15\linewidth}L@{}}
\toprule
\multicolumn{2}{@{}l}{\textbf{Video-prompt alignment}} \\
Confirmed & the quote is literal contract text (checked by program), the requirement applies to the issue's shots, and a supporting frame of those shots contradicts it \\
Rejected & the contract does not state the claim (including requirements only the original prompt states); the requirement does not apply to these shots; the frames satisfy the words (with a refuting frame); the shot satisfies the words and only differs from another shot \\
Uncertain & the frames cannot decide; the missing evidence is named \\
\midrule
\multicolumn{2}{@{}l}{\textbf{Intra-video alignment}} \\
Confirmed & the observations should match (same entity, place, or moment, or one continuous take), no normal cause explains the difference (the causes checked are stated), and the evidence shows both sides \\
Rejected & the frames show no difference (with a refuting frame); a named normal cause explains it; the observations are not the same entity, place, or moment; the video is consistent with itself and only contradicts the text \\
Uncertain & the frames cannot decide; the missing evidence is named \\
\bottomrule
\end{tabularx}
\end{table}

The runner then computes the deduction $D_u = w(\mathrm{sev}_u)\,\gamma(\mathrm{conf}_u)$ and the dimension score $s^d_i = 100\exp\!\big(-\sum_{u} D_u/(100N_i)\big)$ exactly as in Sec.~3.3.4.
Here $N_i$ is the number of shots, so $\sum_u D_u/N_i$ is the deduction per shot.
The total deduction is thus normalized by the number of shots, for the reasons given in Sec.~\ref{sec:length}.
A full deduction of 100 points per shot gives $100/e \approx 36.8$.
A single confirmed moderate issue with high confidence gives 93.94, 95.92, and 96.92 in an episode of 4, 6, and 8 shots, respectively.
The runner rounds the scores to two decimals and averages the three unrounded dimension scores into the UEC score.
The score of a model is the mean over episodes.

The UEC score is the overall score of the benchmark over its three dimensions.
Video-prompt alignment covers the requirements of the contract, and the two intra-video dimensions cover every entity in the video, specified or unspecified.
These two dimensions are where unspecified entities enter the score (Sec.~\ref{sec:uec-scope}).

\begin{table}[t]
\caption{\textbf{Severity levels} given to the severity agents, with the penalty $w$ of each level.}
\label{tab:severity}
\centering
\scriptsize
\begin{tabularx}{\linewidth}{@{}P{0.13\linewidth}P{0.04\linewidth}L@{}}
\toprule
Severity & $w$ & Meaning \\
\midrule
\multicolumn{3}{@{}l}{\textbf{Video-prompt alignment}} \\
Minor & 10 & small, local mismatch; the main identity, action, or relation remains recognizable \\
Moderate & 25 & clear mismatch limited to one detail, subject, relation, or a short part of a shot \\
Major & 50 & a central required action, entity, or relationship fails, or a described entity is substantially wrong across shots \\
Critical & 80 & defeats a core premise of the episode or affects several important shots or entities \\
\midrule
\multicolumn{3}{@{}l}{\textbf{Intra-video alignment}} \\
Minor & 10 & small, local inconsistency; the entity, place, or moment remains recognizable as the same \\
Moderate & 25 & clear inconsistency limited to one detail, subject, relation, or a short part of a shot \\
Major & 50 & a recurring entity or place changes substantially (another face, outfit, or building), or a central element of a shot is corrupted \\
Critical & 80 & breaks a core premise or affects several important shots or entities across the episode \\
\bottomrule
\end{tabularx}
\end{table}

\subsection{Validation-repair loop}
\label{sec:repair}

\paragraph{Repair.}
After every model call, a program checks the output against its schema and every deterministic rule of the stage.
For example, every requirement must be answered once, quotes must be literal, and every candidate detection must be assigned once.
Likewise, cited frames and entity keys must exist, an inter-shot issue must have evidence from two shots, and a confirmed verdict must cite a supporting frame.
All violations go back to the model together with its previous output.
The model is instructed to fix every listed problem in one pass and to keep everything else unchanged.
The instruction also tells it to use only identifiers that exist or that the task allows it to create, and never to invent evidence to satisfy the checker.
The corrected output is checked again from scratch, for at most eight repair rounds.
Shot clustering instead makes up to 12 fresh calls per shot, each listing the problems of the previous answer.
From the third call on, the program assigns any candidate that the model leaves out to the rejections.
This assignment happened once in the 1,000 videos of the first five models and in none of the 120 human-study videos.
Transient transport errors (rate limits, server errors, dropped connections, cut-off replies) are retried up to three times.

\paragraph{How often repair acts.}
In the 120 human-study videos, 229 of 1,220 shot clusterings needed a correction, and 15 needed two or more.
In video-prompt detection on the same videos (run 2), the first answer failed the checks for 46 videos.
The failures came mostly from entities that the ledger locates but that were answered not verifiable (100 errors) and from the absence rule (120 errors).
Of the 218 requirements about ledger-located entities that were first answered not verifiable, 105 became satisfied after repair, 37 violated, and 76 stayed not verifiable with a stated reason.

\begin{table}[t]
\caption{\textbf{Agreement of three complete runs} on 120 videos. ICC(2,1): absolute agreement with a 95\% interval from an episode-level bootstrap. SD: per-video standard deviation over the runs, mean over videos. Range\,$>$10: share of videos whose three scores span more than 10 points.}
\label{tab:repeats}
\centering
\scriptsize
\setlength{\tabcolsep}{3pt}
\begin{tabular*}{\linewidth}{@{\extracolsep{\fill}}lcccc@{}}
\toprule
Dimension & ICC(2,1) & Means (runs 0/1/2) & SD & Range\,$>$10 \\
\midrule
Video-prompt & 0.92 [0.88, 0.95] & 64.8 / 64.3 / 64.5 & 6.44 & 45.8\% \\
Intra-shot & 0.52 [0.30, 0.65] & 97.4 / 96.2 / 97.2 & 2.28 & 15.8\% \\
Inter-shot & 0.86 [0.76, 0.91] & 85.7 / 85.4 / 86.3 & 4.76 & 38.3\% \\
Intra-video (mean) & 0.83 [0.73, 0.89] & 91.5 / 90.8 / 91.7 & 2.78 & 10.0\% \\
UEC & 0.91 [0.87, 0.93] & 82.6 / 81.9 / 82.7 & 2.97 & 13.3\% \\
\bottomrule
\end{tabular*}
\end{table}

\subsection{Calls and time}
\label{sec:cost}

Over the 1,000 videos of the first five models, an episode takes 76.4 model calls on average, including repair rounds and retries.
The formula of Table~\ref{tab:calls} gives the same value at the mean $N=5$ and $K=14.4$, because the added repair calls and the skipped calls nearly cancel.
In the two repeated runs of Sec.~\ref{sec:repeats}, the pipeline took 700 and 704\,s per video on average, with up to three videos sharing one GPU and the WeDetect detections reused.
These averages are not the time of an isolated run.
At the GPT-6 Luna list price, an evaluation costs on the order of USD 0.1 per video.

\section{Reliability, Sensitivity, and the Role of Unspecified Entities}
\label{sec:reliability}

\subsection{Repeated runs and other evaluation models}
\label{sec:repeats}

On the 120 human-study videos (Sec.~\ref{sec:human}), the default pipeline was run twice more (runs 1 and 2) with the code of the main run (run 0).
Every model call was made afresh.
Only the WeDetect detections of run 0 were reused, because the detector makes no model call and, for most videos, a rerun would have used a different device.
Table~\ref{tab:repeats} gives the agreement of the three runs.
ICC(2,1) is 0.92 for video-prompt alignment, 0.52 for intra-shot and 0.86 for inter-shot consistency, and 0.91 for UEC.
Intra-shot consistency has the lowest ICC.
Most videos score at or near 100 on it, and intra-shot issues are few.
The three runs confirmed 63, 92, and 66 intra-shot issues, against 1,531, 1,592, and 1,511 video-prompt issues and 309, 316, and 302 inter-shot issues.
One intra-shot issue found in one run but not another therefore moves the score of a video by several points.
The five-model order is identical in all three runs for video-prompt alignment, inter-shot consistency, and the mean of the two intra-video dimensions.
The UEC order changes in one pair.
HoloCine and Wan2.2 swap places in runs 1 and 2 (Kendall's $\tau$ 0.8 against run 0), and their run-0 means differ by 0.63 points (HoloCine 87.22, Wan2.2 87.85).

\paragraph{Other evaluation models.}
The two medium-effort rows of Table~5 come from reruns of the evaluation stages (detection, the unspecified-object check, verification, and scoring) on the 120 videos.
These reruns used GPT-6 Luna and Gemini~3.8 Flash, both at medium effort, and reused the entity ledgers of run 0.
Prompts, rules, and schemas were unchanged, except for one change of form in the schemas for Gemini.
The default form is an object with one required field per requirement or issue ID.
The Gemini API rejected it for long requirement lists, so the Gemini video-prompt answers and verdicts were requested as arrays of rows keyed by these IDs.
The Gemini checklist calls also used explicit context caching.
Both models are stricter than the default.
Their mean UEC is 77.0 (Luna medium) and 70.2 (Gemini) against 82.6, and they confirm 2,132 and 2,624 video-prompt issues against 1,531.
Gemini orders the five models on UEC exactly as run 0 does ($\tau$ 1.0), and Luna medium swaps HoloCine and Wan2.2 ($\tau$ 0.8).
Per video, their UEC scores correlate with those of run 0 at Spearman 0.85 (Luna medium) and 0.72 (Gemini), against 0.85 and 0.87 for the two repeated Luna-low runs.
The main paper states that the agreement of UEC-Bench with the raters does not depend on a specific evaluation model or reasoning effort.
This statement refers to the agreement of Table~5 and to the model order.

\subsection{Intervals for Table~4}
\label{sec:ci}

Table~\ref{tab:ci} repeats Table~4 with 95\% intervals from an episode-level bootstrap on the run of Table~4.
The 200 episodes were resampled with replacement 10,000 times with a fixed seed, and every model and dimension used the same resamples.
Since each model has one video per episode, resampling episodes resamples the videos of that model.
The mean score of each model was recomputed on every resample, and the interval runs from the 2.5th to the 97.5th percentile of these means.
The point estimates are the means of the stored per-video scores and reproduce Table~4.
The computation makes no model call.

\begin{table*}[t]
\caption{\textbf{Table~4 with 95\% intervals.} Mean over the 200 episodes of each model and, in brackets, the 95\% percentile interval of an episode-level bootstrap (10,000 resamples).}
\label{tab:ci}
\renewcommand{\arraystretch}{1.1}
\centering
\small
\begin{tabular*}{\textwidth}{@{\extracolsep{\fill}}lcccc@{}}
\toprule
 & & \multicolumn{2}{c}{Intra-video} & \\
\cmidrule(lr){3-4}
Model & Video-prompt & Intra-shot & Inter-shot & UEC (avg.) \\
\midrule
Wan2.1-T2V-1.3B & 64.2 [61.6, 66.8] & 96.3 [95.4, 97.0] & 71.4 [69.4, 73.3] & 77.3 [76.1, 78.5] \\
Wan2.2-T2V-A14B & 82.6 [80.7, 84.3] & 98.0 [97.4, 98.5] & 75.7 [73.9, 77.5] & 85.4 [84.5, 86.3] \\
LongLive-1.3B & 37.3 [34.4, 40.3] & 87.1 [85.4, 88.8] & 88.6 [87.1, 90.1] & 71.0 [69.5, 72.4] \\
\midrule
CineTrans-DiT-1.3B & 21.7 [19.3, 24.2] & 95.0 [94.3, 95.7] & 85.5 [83.6, 87.3] & 67.4 [66.4, 68.4] \\
MultiShotMaster-1.3B & 55.1 [51.9, 58.4] & 97.4 [96.8, 98.0] & 90.4 [89.0, 91.7] & 81.0 [79.7, 82.2] \\
HoloCine-A14B & 65.7 [62.3, 69.0] & 96.8 [96.1, 97.5] & 96.4 [95.7, 97.1] & 86.3 [85.1, 87.5] \\
\midrule
ShotStream-1.3B & 44.7 [41.8, 47.6] & 98.7 [98.2, 99.1] & 90.6 [89.3, 91.8] & 78.0 [76.8, 79.1] \\
StoryMem-A14B & 84.0 [82.0, 85.8] & 97.1 [96.4, 97.6] & 92.4 [91.4, 93.4] & 91.2 [90.4, 92.0] \\
\bottomrule
\end{tabular*}
\end{table*}

\subsection{Scoring hyperparameters}
\label{sec:sensitivity}

All per-video scores of the first five models were recomputed under alternative scoring parameters from the stored issues of the main run, with zero model calls.
The recomputation covers all 1,000 videos and the 120 human-study videos.
With the default values, it reproduces every stored score exactly.
Table~\ref{tab:sensitivity} lists the alternatives and how often the five-model UEC order is kept.

The order is insensitive to the uncertain factor, varied from 0 to 1, and to the confidence factors.
Across the values of the uncertain factor, the mean absolute change of the per-video UEC score is at most 0.65.
Five of the six alternative severity scales keep the order on all 1,000 videos, and four of six keep it on the subset.
Every change is a swap of HoloCine and Wan2.2, and the alternative dimension weights of Table~\ref{tab:sensitivity} change only the same pair.
Outside the weight family, the per-video UEC score keeps a Spearman correlation of 0.98 to 1.00 with the default.

The alternatives were also combined in a full factorial of 504 combinations.
On both sets, the minimum Kendall's $\tau$ against the default order is 0.80.
The exact order is kept by 217 combinations on all videos and by 346 on the subset.
The only swaps are HoloCine with Wan2.2 (270 and 158 combinations) and, on all videos, Wan2.1 with CineTrans-DiT (17).
The intra-video order (mean of intra- and inter-shot) is unchanged in all 504 combinations.

The defaults were set by design and were not fitted to the human ratings.
The severity penalties come from a rubric used since the first evaluator version, and the confidence and uncertain factors were fixed before the human study.
The UEC score is the plain mean of the three dimensions.
Unspecified entities enter it through the two consistency dimensions, and Table~4 reports each dimension and the unspecified-entity measures separately (Sec.~\ref{sec:uec-scope}).

\begin{table}[t]
\caption{\textbf{Sensitivity of the five-model order by UEC} to the scoring parameters, one family changed at a time (all 1,000 videos / 120-video subset).}
\label{tab:sensitivity}
\centering
\scriptsize
\begin{tabularx}{\linewidth}{@{}P{0.27\linewidth}Ll@{}}
\toprule
Family (default) & Alternatives & Order kept \\
\midrule
Severity penalties\newline (10/25/50/80) & 25/25/25/25; 20/40/60/80; 10/20/40/80; 5/15/40/100; default $\times$0.5; default $\times$2 & 5 of 6 / 4 of 6 \\
Uncertain factor\newline (0.25) & 0; 0.125; 0.5; 0.75; 1 & 5 of 5 / 5 of 5 \\
Confidence factors\newline (1/0.75/0.5) & 1/1/1; 1/0.5/0.25 & 2 of 2 / 2 of 2 \\
Dimension weights\newline (1/3 each) & 1/2,1/4,1/4; 1/5,2/5,2/5; 3/5,1/5,1/5 & 1 of 3 / 2 of 3 \\
\bottomrule
\end{tabularx}
\end{table}

\subsection{Episode length, persistent problems, and unobservable content}
\label{sec:length}

\paragraph{Why the deduction is normalized by the number of shots.}
The score uses the deduction per shot, the total deduction of an episode divided by its number of shots (Sec.~\ref{supp:sec:scoring}).
An earlier scoring version multiplied $(1-D_u/100)$ over issues.
In a development run over the same 1,000 videos, this rule had two problems.
It saturated: 268 videos had a video-prompt score below 1, so further issues no longer changed their scores.
It also penalized length: mean UEC was 57.65, 51.43, and 44.67 for episodes of 4, 6, and 8 shots.
Three parameter-free per-shot forms were compared: the per-shot geometric mean of the product, the exponential of the per-shot deduction, and $100/(1+x)$ with $x=\sum_u D_u/(100N_i)$.
All three gave the same model order.
The exponential was chosen because it is the simplest and its per-video score distribution was closest to that of an earlier evaluator version.
The model order was not a selection criterion, and no human rating was used.
In the main run, mean UEC is 81.65, 81.32, and 81.10 for episodes of 4, 6, and 8 shots (600, 300, and 100 videos).
Confirmed issues per video, however, grow from 11.71 to 20.56 and 30.68.%
\footnote{Computed on the main run before the correction described in the footnote of Sec.~\ref{sec:uoc}, which changed the scores of 26 of the 1,000 videos.}

\paragraph{Persistent problems.}
An inconsistency that persists over several shots is one intra-video issue covering all of them.
The verifier merges repeated reports of it, so it deducts once.
The extent of such an issue enters through its severity.
The rubric reserves critical for an inconsistency affecting several important shots or entities across the episode, and the severity agent judges each issue at the full extent described.
With the normalization by the number of shots, the score measures the number of distinct problems per shot.
The number of shots an issue spans enters only through its severity, so an issue with a fixed deduction weighs less in a longer episode.
For example, a confirmed major issue with high confidence ($D=50$) gives an inter-shot score of 88.25 in a four-shot episode and 93.94 in an eight-shot episode.
Video-prompt alignment differs, because every failed requirement is a separate issue.
A required entity missing in several shots fails in each of them, together with every requirement that needs it.
In the main run, confirmed intra-video issues per video grow more slowly than the shot count.
They number 3.10, 4.08, and 4.91 for 4, 6, and 8 shots, of which 2.53, 3.46, and 3.91 are inter-shot issues.
The intra-video deduction per shot is therefore somewhat smaller in longer episodes, while the video-prompt issues grow roughly in proportion (8.61, 16.48, and 25.77).

\paragraph{Unobservable content.}
A requirement that the frames cannot decide is answered not verifiable and creates no issue.
They are 4.6\% of all answers in the 120 human-study videos.
The intra-video agent likewise reports a change only when the frames show it.
Content that is never visible therefore lowers no score.
A contract entity that never appears, however, is a presence failure in each shot that lists it.
A candidate that the verifier cannot decide is kept as uncertain and deducts a quarter of its severity penalty.
In run 0 on the same videos, the confirmed and uncertain issues were 1,693 video-prompt, 66 intra-shot, and 361 inter-shot issues, of which 162, 3, and 52 were uncertain.

\subsection{Entity density and coverage}
\label{sec:density}

Table~\ref{tab:density} gives, per model, the unspecified entities and the confirmed issues behind Table~4.
The number of unspecified entities depends on what each model renders and ranges from 960 for CineTrans-DiT to 2,464 for Wan2.2.
CineTrans-DiT, at the low end, often renders nearly static, sparse shots.
Its median mean absolute difference between consecutive keyframes is 2.65 on a 0--255 scale, against 6.89 to 14.85 for the other four of the first five models.

A model that renders fewer unprompted objects has fewer chances of an unspecified inconsistency.
For this reason, Table~4 also reports the entity-normalized columns (share of entities with issues).
Omitted required content, in contrast, is penalized by video-prompt alignment, where CineTrans-DiT scores 21.7.
Static or repeated content fails the presence, motion, and cut requirements of the contract.
CineTrans-DiT has 3,132 confirmed presence issues and 1,412 action issues, and LongLive, which cannot cut, has 121 confirmed missing cuts (Table~\ref{tab:subcat-counts}).
Intra-video alignment alone does not penalize such output, and the UEC score combines it with video-prompt alignment.

\begin{table}[t]
\caption{\textbf{Unspecified entities and confirmed issues per model} (200 episodes each). Unsp.: unspecified entities. VV: confirmed intra-video issues. (unsp.): those of them with an unspecified subject. VP: confirmed video-prompt issues.}
\label{tab:density}
\centering
\scriptsize
\setlength{\tabcolsep}{3pt}
\begin{tabular*}{\linewidth}{@{\extracolsep{\fill}}lrrrr@{}}
\toprule
Model & Unsp. & VV & (unsp.) & VP \\
\midrule
Wan2.1 & 2,095 & 1,206 & 293 & 1,715 \\
Wan2.2 & 2,464 & 1,015 & 280 & 710 \\
LongLive & 1,762 & 853 & 279 & 3,925 \\
CineTrans-DiT & 960 & 674 & 179 & 7,737 \\
MultiShotMaster & 2,349 & 446 & 126 & 2,392 \\
HoloCine & 1,426 & 259 & 65 & 1,848 \\
ShotStream & 1,288 & 381 & 133 & 3,160 \\
StoryMem & 2,069 & 416 & 159 & 674 \\
\midrule
All eight & 14,413 & 5,250 & 1,514 & 22,161 \\
\bottomrule
\end{tabular*}
\end{table}

\subsection{Specified versus unspecified entities}
\label{sec:denominators}

Table~\ref{tab:cross-split} splits the confirmed inter-shot issues of each model by subject label.
An issue counts as unspecified when at least one of its subjects is unspecified.
Issues without a subject entity, such as a lighting change, are listed separately.
Relative to its base model, every multi-shot method has fewer inter-shot issues of both kinds, and the reduction is larger for specified subjects.
Against Wan2.2, specified issues fall by 74.5\% and unspecified ones by 51.3\% for StoryMem, and by 85.6\% and 80.6\% for HoloCine.
Against Wan2.1, they fall by 55.8\% and 42.6\% for CineTrans-DiT, 71.2\% and 59.3\% for MultiShotMaster, and 72.8\% and 57.8\% for ShotStream.

\begin{table}[t]
\caption{\textbf{Confirmed inter-shot issues by subject type} (200 episodes per model). Reduction: relative to the base model (Wan2.1 for CineTrans-DiT, MultiShotMaster, and ShotStream; Wan2.2 for HoloCine and StoryMem).}
\label{tab:cross-split}
\centering
\scriptsize
\setlength{\tabcolsep}{2.5pt}
\begin{tabular*}{\linewidth}{@{\extracolsep{\fill}}lrrrrr@{}}
\toprule
 & \multicolumn{3}{c}{Issues} & \multicolumn{2}{c}{Reduction} \\
\cmidrule(lr){2-4}\cmidrule(l){5-6}
Model & Spec. & Unsp. & None & Spec. & Unsp. \\
\midrule
Wan2.1 & 740 & 263 & 67 & -- & -- \\
Wan2.2 & 596 & 263 & 77 & -- & -- \\
LongLive & 215 & 174 & 3 & -- & -- \\
CineTrans-DiT & 327 & 151 & 5 & 55.8\% & 42.6\% \\
MultiShotMaster & 213 & 107 & 25 & 71.2\% & 59.3\% \\
HoloCine & 86 & 51 & 8 & 85.6\% & 80.6\% \\
ShotStream & 201 & 111 & 10 & 72.8\% & 57.8\% \\
StoryMem & 152 & 128 & 27 & 74.5\% & 51.3\% \\
\midrule
All eight & 2,530 & 1,248 & 222 & & \\
\bottomrule
\end{tabular*}
\end{table}

\subsection{Contribution of the unspecified-object check}
\label{sec:ablation}

\paragraph{Examination of the two groups.}
Specified entities receive a per-requirement examination from the contract.
Video-prompt alignment answers every requirement of every specified entity in every shot where it applies (Sec.~\ref{sec:vp}).
Discovery searches for every contract entity expected in a shot, and contract IDs link identities across shots (Sec.~\ref{supp:sec:discovery}).
The checklist calls cover the consistency of all entities.
The unspecified-object check is part of the core purpose of the benchmark.
It gives the objects that the prompt leaves unnamed a dedicated per-object examination of the same kind.
The check lists these objects shot by shot, compares each within its shot and with every other shot of its location, and verifies every suspect (Sec.~\ref{sec:uoc}).
It is thus a coverage step, and its contribution is what it finds.

\paragraph{Ablation.}
To measure this contribution, the evaluator was also run without the check on the 1,000 videos of the first five models.
Both runs share the entity ledgers, the video-prompt results, and the checklist candidates.
The full pipeline adds the check and verifies the merged candidates (Table~\ref{tab:ablation}).
The check raises the confirmed intra-video issues with an unspecified subject from 319 to 976, and their share from 10.5\% to 27.3\%.
It lowers the mean inter-shot score from 85.85 to 84.29.

\begin{table}[t]
\caption{\textbf{Evaluator with and without the unspecified-object check} (1,000 videos of five models).}
\label{tab:ablation}
\centering
\scriptsize
\setlength{\tabcolsep}{3pt}
\begin{tabular*}{\linewidth}{@{\extracolsep{\fill}}lcc@{}}
\toprule
 & Without check & With check \\
\midrule
Video-prompt / intra-shot / inter-shot & 63.64 / 96.56 / 85.85 & 63.64 / 96.62 / 84.29 \\
UEC & 82.01 & 81.52 \\
Confirmed intra-video issues & 3,039 & 3,570 \\
\quad with an unspecified subject & 319 (10.5\%) & 976 (27.3\%) \\
Unspecified entities & 8,417 (27.5\%) & 9,014 (28.9\%) \\
\bottomrule
\end{tabular*}
\end{table}

\subsection{Specified-only proxy}
\label{sec:proxy}

UEC-Bench reports the consistency failures of every entity in the video.
A failure of an unprompted object, such as a garment changing between shots, is a failure of the video whether or not it changes an aggregate ranking.
The value of evaluating unspecified entities therefore does not depend on changing model rankings or on raising the agreement with an overall human rating.
The main paper makes no claim that evaluating unspecified entities raises this agreement.
The specified-only proxy below describes how much these failures contribute to the scores.
It is an approximation that drops every confirmed or uncertain intra-video issue with an unspecified subject and recomputes the scores.

\paragraph{The proxy.}
The proxy approximates a prompt-centric evaluation that would use the same pipeline but ignore unspecified entities.
It is computed from the stored issues with zero model calls.
Every confirmed or uncertain intra-video issue whose subjects include an unspecified entity is dropped, and the scores are recomputed by the same rule (Table~\ref{tab:proxy}).
Because the proxy also drops issues that name prompt entities together with unspecified ones, it removes somewhat more than a prompt-centric evaluation would necessarily miss.
On the run of Table~4, the proxy raises the intra-video scores of every model.
Over the eight models, the mean inter-shot score rises from 86.37 to 89.34, and the mean UEC from 79.69 to 80.92.
The rise is largest for the three baselines (inter-shot $+4.45$, $+4.46$, and $+4.09$ for Wan2.1, Wan2.2, and LongLive) and smallest for HoloCine ($+0.98$).
Two pairs change places in the UEC order.
Wan2.2 moves ahead of HoloCine, 87.03 against 86.75 (full evaluation: 85.42 against 86.33).
Wan2.1 moves ahead of ShotStream, a multi-shot method built on it, 78.99 against 78.96 (full evaluation: 77.28 against 77.97).
In the inter-shot order, LongLive moves ahead of MultiShotMaster, 92.69 against 92.53 (full evaluation: 88.60 against 90.37).
The intra-shot order stays the same.

\begin{table}[t]
\caption{\textbf{Full evaluation and specified-only proxy} on the run of Table~4: inter-shot and UEC means of each model (200 videos).}
\label{tab:proxy}
\centering
\scriptsize
\setlength{\tabcolsep}{3pt}
\begin{tabular*}{\linewidth}{@{\extracolsep{\fill}}lcccc@{}}
\toprule
 & \multicolumn{2}{c}{Inter-shot} & \multicolumn{2}{c}{UEC} \\
\cmidrule(lr){2-3}\cmidrule(l){4-5}
Model & Full & Proxy & Full & Proxy \\
\midrule
Wan2.1 & 71.39 & 75.85 & 77.28 & 78.99 \\
Wan2.2 & 75.71 & 80.18 & 85.42 & 87.03 \\
LongLive & 88.60 & 92.69 & 70.99 & 73.13 \\
CineTrans-DiT & 85.48 & 88.40 & 67.39 & 68.55 \\
MultiShotMaster & 90.37 & 92.53 & 80.98 & 81.86 \\
HoloCine & 96.44 & 97.42 & 86.33 & 86.75 \\
ShotStream & 90.56 & 92.98 & 77.97 & 78.96 \\
StoryMem & 92.43 & 94.67 & 91.16 & 92.12 \\
\midrule
All eight & 86.37 & 89.34 & 79.69 & 80.92 \\
\bottomrule
\end{tabular*}
\end{table}

\begin{figure}[t]
\centering
\includegraphics[width=\linewidth]{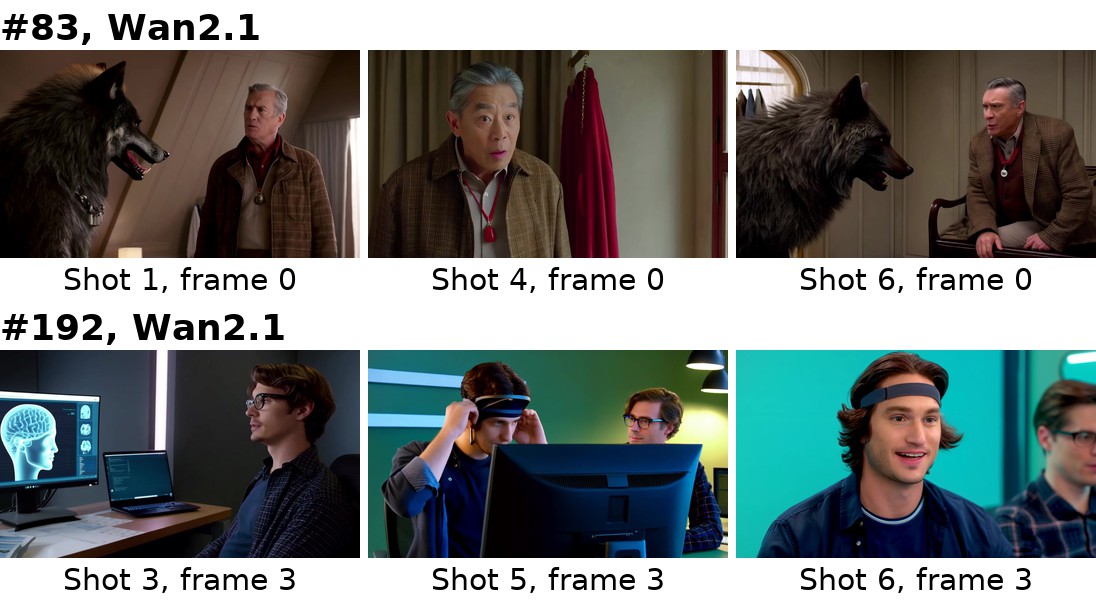}
\caption{\textbf{Keyframes cited by the issue the specified-only proxy drops} in the Wan2.1 videos of episodes \#83 and \#192 (downscaled). \#83: the room behind the man is a dark brown interior with pale paneling and beige and gray garments in Shots 1, 2, and 6 (Shot 2 not shown), and a pale plain wall with a red hanging garment in Shot~4. The different face of the man in Shot~4 is a separate confirmed identity issue of a prompt-named character, which the proxy keeps. \#192: the walls of one workspace are dark gray, green, and turquoise in Shots 3, 5, and 6. In both episodes, the raters and the full evaluation rank the Wan2.1 video below the video of another model. After dropping these issues, the proxy ranks it above.}
\label{fig:proxy}
\end{figure}

\paragraph{The proxy and the raters.}
On the 120 human-study videos, with the UEC-Bench run of Table~5, the proxy changes the score of 57 videos.
The agreement of the proxy with the overall human rating is $\rho$ 0.742 and $\tau_b$ 0.667, against 0.746 and 0.675 for the full evaluation (Table~5).
A within-episode pair has a clear human preference when the mean overall ratings of its two videos differ by at least 0.5 (Sec.~\ref{sec:human-results}).
Of the 176 such pairs, the proxy changes the order of 5.
In 3 of them, the full evaluation orders the pair as the raters do and the proxy does not.
In the other 2, the reverse holds.
Figure~\ref{fig:proxy} shows the two clearest cases of the first kind.
In episode \#83, the raters and the full evaluation prefer the HoloCine video to the Wan2.1 video (UEC 87.28 against 86.11).
The deciding issue is an inter-shot change of the room in the Wan2.1 video.
Shots 1, 2, and 6 show a dark brown interior with pale paneling and beige and gray garments.
In Shot~4, the man instead stands before a pale plain wall with a red hanging garment.
Besides two prompt-named elements, the subjects of this issue include the unprompted wall and red garment.
The proxy therefore drops the issue and ranks Wan2.1 first (87.87).
In episode \#192, the raters and the full evaluation prefer the Wan2.2 video to the Wan2.1 video (89.87 against 89.40).
In the Wan2.1 video, the walls of one workspace change across Shots 3, 5, and 6.
They are dark gray with a narrow vertical light strip, green with a horizontal light-blue strip, and bright turquoise, respectively.
The subjects of the issue reporting this change include an unprompted turquoise wall and light strip.
After the proxy drops this issue, Wan2.1 scores 91.54.
In the third pair of this kind (\#134), the dropped issue is an uncertain change of a man's jacket in the HoloCine video.
The full scores of this pair differ by only 0.02.
In the two pairs of the other kind, the dropped issue concerns only unprompted items in the video that the raters prefer.
In \#104 (Wan2.2), a woman's earrings change from studs to hoops, and in \#175 (Wan2.2), sofa cushions disappear between shots.
Changes involving unprompted objects can thus decide how a video ranks, for the raters and for UEC-Bench.
Over all 120 videos, however, the two versions agree with the raters almost equally.

\paragraph{What this comparison can show.}
The comparison has little power to separate the two versions, for two reasons.
First, the overall human rating is the mean of consistency with the text, within shots, and across shots.
The proxy changes only the intra-video part of the score, so the text third of the rating and the video-prompt third of the score are untouched.
Second, only 5 of the 176 pairs with a clear human preference change order between the versions.
These are too few for the split of 3 against 2 to favor either version.
The across-shots question told the raters to count changes of things the text does not mention (Table~\ref{tab:rater-instructions}), so unspecified entities are part of the human target.

\section{Human Study}
\label{sec:human}

\subsection{Protocol}
\label{sec:human-protocol}

\paragraph{Videos.}
The 24 episodes of the study were drawn from 162 of the 200 benchmark episodes.
The other 38 episodes had been inspected during evaluator development before the sample was taken and were excluded.
The stratified random draw with a fixed seed gives 14, 7, and 3 episodes with 4, 6, and 8 shots, respectively.
Of these episodes, 14 come from SF20K and 10 from MuSS, 3 are animated, and each source work contributes at most one episode.
No evaluator score was used for the draw.
Each episode contributes the videos of Wan2.1, Wan2.2, CineTrans-DiT, HoloCine, and StoryMem, giving 120 videos.

\paragraph{Raters and assignment.}
Ten raters took part in total, and each episode was rated by five raters.
Each rated the five videos of the episode together.
The raters were recruited by the authors.
The five videos appear under random labels A--E, and model names, contracts, and evaluator outputs are hidden.
The study had no practice item, and all raters and ratings were kept.
A rater took a median of 5.7 minutes to rate one episode (five videos, three questions each).

\paragraph{Interface.}
For each episode, the interface shows the global and shot prompts in the English original, with a Chinese reference translation.
The English original is the text the evaluator reads.
The interface is in Chinese, and the English text is authoritative.
The videos play one at a time, with a strip that marks the shot boundaries and highlights the text of the current shot.
During the study, the interface gained a keyframe thumbnail and a one-line summary per shot.
The instructions, questions, anchors, videos, and prompts stayed the same.
Some ratings had been given with the earlier interface, and all ratings are pooled.

\paragraph{Instructions.}
Raters answer three questions per video on a five-point scale with verbal anchors (Table~\ref{tab:rater-instructions}).
The questions cover the three aspects of entity consistency that the benchmark scores, and raters judge each aspect separately.
These aspects are consistency with the text (whether the video matches the text description), consistency within shots, and consistency across shots.
The rating of a video on a question is the mean over its raters.
The overall rating of a video is the mean of its three question ratings and is the target of Table~5.

\begin{table}[t]
\caption{\textbf{Rater instructions} (translated from Chinese). General instructions: \emph{Each item has a text description and five videos. The description is the English original with a Chinese reference translation below it; where they differ, the English is authoritative. After watching each video, answer three questions; 1 is worst and 5 is best. Judge the three questions separately; visual appeal and sharpness are not scored.}}
\label{tab:rater-instructions}
\centering
\scriptsize
\begin{tabularx}{\linewidth}{@{}L@{}}
\toprule
\textbf{Matches the text.} Does the video match the text description? \emph{Compare shot by shot with the text: characters, objects, places, appearance, actions, positions, shot scale. What the text does not state does not count.} 5 fully matches; 4 one or two small details do not match; 3 one important element does not match, or several small details do not; 2 several important elements do not match; 1 basically does not match. \\
\midrule
\textbf{Consistent within shots.} Is each shot consistent from beginning to end? \emph{For example, a face turns into someone else, an object changes color or shape, something appears or disappears from nowhere, limbs are malformed, the image flickers. Camera movement and actions the text describes do not count.} 5 no problems; 4 one or two slight flaws; 3 one clear problem; 2 several clear problems, or one severe problem; 1 most shots are unstable. \\
\midrule
\textbf{Consistent across shots.} Are the same people, objects, and settings consistent in different shots? \emph{For example, a changed face or clothes, a changed room layout; things the text does not mention also count. Changes the text asks for (a change of scene or of clothes) do not count.} 5 fully consistent; 4 small differences in details; 3 one clear inconsistency; 2 several clear inconsistencies, or a main character or setting looks replaced; 1 the shots are unrelated. \\
\bottomrule
\end{tabularx}
\end{table}

\paragraph{Statistics.}
Spearman's $\rho$ between a metric and the mean human overall rating is computed over all 120 videos.
Kendall's $\tau_b$ is computed within each episode over its five videos and then averaged over episodes.
No episode has a constant metric or human mean on the overall rating.
All 95\% intervals come from 10,000 bootstrap resamples of whole episodes.
As a reference for the agreement between raters, the overall ratings of each rater are correlated with the left-out mean, that is, the mean of the other raters on the same videos.
These correlations are averaged over raters.
Each metric is correlated with the same left-out means.

\subsection{Baselines}
\label{sec:baselines}

All baselines use the same shot windows, keyframes, and prompts as the evaluator.
\emph{ViCLIP similarity} uses the ViCLIP model of VBench~\cite{huang2024vbench,wang2024internvid} with eight frames per shot.
Its prompt score is the mean over shots of the cosine between the shot video and the shot prompt.
Because the text encoder holds only 32 tokens, each shot prompt is split into sentence pieces that fit, and their cosines are averaged.
Its across-shot score is the mean cosine between shot videos over the shot pairs that share at least one contract character and one contract location.
These are 181 of the 273 shot pairs of the 24 episodes.
Its overall score is the mean of the two similarities, each z-scored over the 120 videos.
\emph{VBench consistency} averages the VBench subject consistency (DINO features) and background consistency (CLIP features~\cite{radford2021learning}) per shot and then over shots.
This average is also its overall score.
The \emph{direct VLM rating} makes one GPT-6 Luna call (low effort) per video.
The call receives the same labeled keyframes as the evaluator, the original prompt, and the rater instructions and questions verbatim.
It receives no contract and no entity ledger.
For each question, it writes a short reason and an integer from 1 to 5, and its overall score is the mean of these three integers.

\subsection{Results}
\label{sec:human-results}

\paragraph{Agreement with the overall rating.}
Table~\ref{tab:human-overall} repeats Table~5 with 95\% intervals.
In the leave-one-rater-out reference, a single rater correlates with the mean of the others at $\rho$ 0.77 on the overall rating.
UEC-Bench at low effort (Table~5) reaches $\rho$ 0.73 against these left-out means.

\begin{table}[t]
\caption{\textbf{Table 5 with 95\% intervals}: Spearman's $\rho$ over all videos and within-episode Kendall's $\tau_b$ with the overall human rating. Every row is the run reported in Table~5. The medium-effort rows rerun the evaluation stages on main-run ledgers. For ViCLIP similarity, the overall value is the mean of its two z-scored similarities. For VBench consistency, it is the consistency score.}
\label{tab:human-overall}
\centering
\scriptsize
\setlength{\tabcolsep}{1.5pt}
\begin{tabular*}{\linewidth}{@{\extracolsep{\fill}}lcc@{}}
\toprule
Metric & $\rho$ & $\tau_b$ \\
\midrule
UEC-Bench, Luna low & 0.75 [0.68, 0.81] & 0.68 [0.58, 0.76] \\
UEC-Bench, Luna medium & 0.76 [0.69, 0.82] & 0.67 [0.58, 0.75] \\
UEC-Bench, Gemini medium & 0.75 [0.69, 0.80] & 0.64 [0.55, 0.73] \\
VBench consistency & $-$0.25 [$-$0.41, $-$0.12] & $-$0.38 [$-$0.51, $-$0.23] \\
ViCLIP similarity & 0.15 [0.03, 0.28] & 0.17 [0.03, 0.30] \\
Direct VLM, Luna low & 0.68 [0.55, 0.78] & 0.49 [0.38, 0.59] \\
\bottomrule
\end{tabular*}
\end{table}

\paragraph{Ties.}
Ties are counted over the 176 within-episode pairs of videos whose mean overall human ratings differ by at least 0.5.
The direct rating of Table~5 ties 25 of these pairs, because its overall score takes only 9 distinct values over the 120 videos.
UEC-Bench, whose overall score takes 110 distinct values, ties none of them.
Among the untied pairs, the direct rating orders 0.86 of them as the raters do, and UEC-Bench orders 0.93.
The direct rating receives the raters' own instructions and anchors, and its overall score is the mean of three integers from 1 to 5.

\subsection{Independence from evaluator development}
\label{sec:independence}

The human ratings were not used to set any evaluator prompt, rule, threshold, or scoring parameter.
The scoring parameters predate the study (Sec.~\ref{sec:sensitivity}).
The per-shot form of the score was chosen without ratings, from the score distribution of all 1,000 videos including the 120 rated ones (Sec.~\ref{sec:length}).
The checklist items were drafted from problem types that earlier evaluator versions reported across the 200 episodes.
Some development tests after the draw used randomly chosen videos of the 200 episodes, and a few of these videos are among the rated ones.
In these tests, the authors inspected evaluator outputs, not ratings.

\section{Examples and Failure Types}
\label{sec:examples}

\subsection{From candidates to a score: one episode}
\label{sec:trace}

\begin{figure*}[t]
\centering
\includegraphics[width=\textwidth]{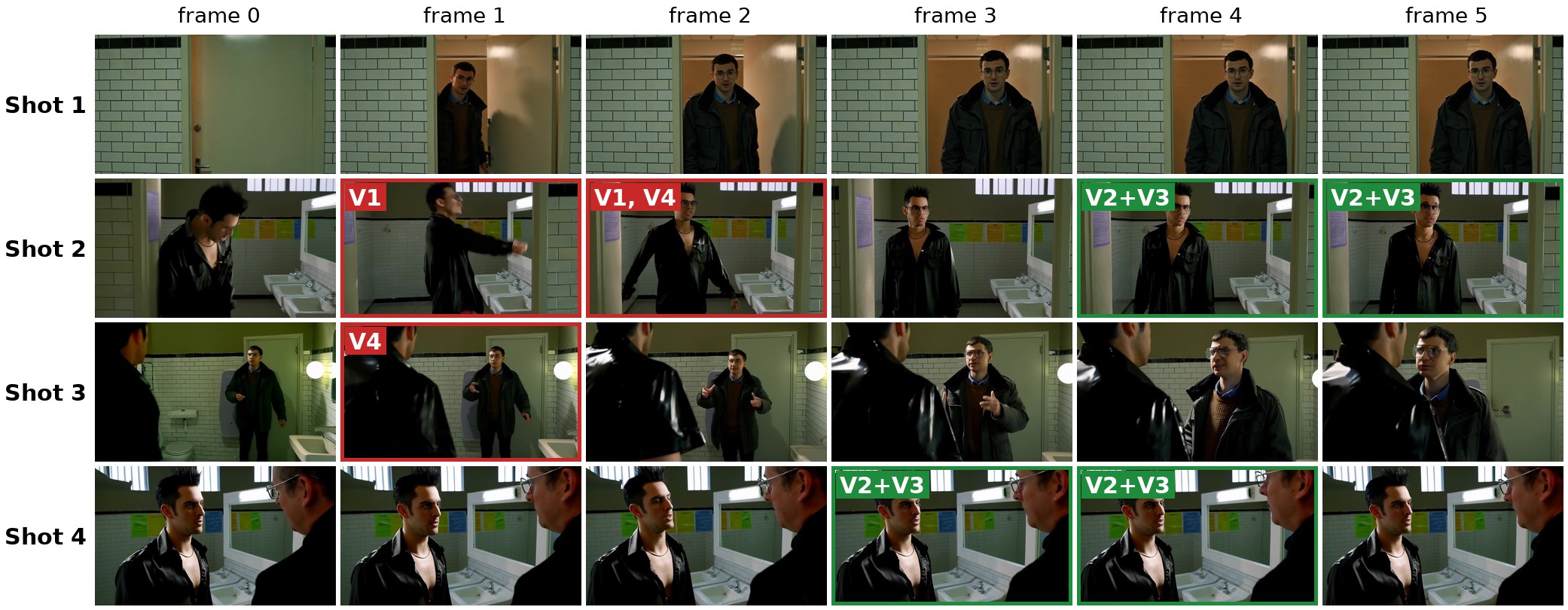}
\caption{\textbf{The 24 keyframes of the StoryMem video for episode \#101}, as attached to the detection and verification calls (downscaled). Outlines mark the frames cited by the candidates of Table~\ref{tab:trace}. Green marks the confirmed issue V2+V3 (glasses worn in Shot 2, absent in Shot 4), and red marks the rejected candidates V1 (glasses during the turn in Shot 2) and V4 (trousers of two different men).}
\label{fig:trace}
\end{figure*}

Table~\ref{tab:trace} follows every intra-video candidate of StoryMem on episode \#101 (Figure~4) through verification and scoring.
Figure~\ref{fig:trace} shows the 24 keyframes that the agents saw.
The checklist (item X3) and the unspecified-object check report the same glasses change independently.
The first verification call merges the two reports, the second confirms the merged issue, and the other two candidates are rejected.
The confirmed issue is rated moderate and deducts $D=25\times1=25$, which gives the inter-shot score $100\exp(-25/400)=93.94$ of Figure~4.
Its subject, the glasses, is an unspecified ledger entity, so the issue counts toward the unspecified share of Table~4.

\begin{table*}[t]
\caption{\textbf{Trace of the intra-video candidates of StoryMem on episode \#101} (translated). DS man: the Dark-shirted man (contract entity 1). Call 1: keep, split, or merge and rewrite; call 2: verdict.}
\label{tab:trace}
\centering
\scriptsize
\begin{tabularx}{\textwidth}{@{}P{0.13\textwidth}>{\raggedright\arraybackslash\hsize=0.8\hsize}X>{\raggedright\arraybackslash\hsize=1.2\hsize}X@{}}
\toprule
Candidate (source; scope; subcategory) & Candidate statement from detection & Verification and scoring \\
\midrule
V1 (checklist W5; intra-shot; object\_instability) & In Shot 2 the glasses are not visible in the first two keyframes and appear in keyframe 2, with no action of putting them on. & Call 1: kept, clarified (the change is between frames 1 and 2, on the man's face). Call 2: rejected, normal change (occlusion or framing), medium confidence: in frame 1 the man is turning and his face is in profile and blurred, so the eye region cannot show that the glasses are absent; the turn explains the change in visibility. No deduction. \\
\midrule
V2 (checklist X3; inter-shot; character\_appearance) & The DS man wears large-framed glasses in Shot 2; in Shot 4 the same man wears none; no action of taking them off or other explanation. & Call 1: merged with V3 into one issue (next row). \\
\midrule
V3 (unspecified-object check; inter-shot; object\_identity) & The object is the large-framed glasses on the DS man's face in Shot 2. The white-framed mirror, sinks, and high windows show the same sink area in both shots; in Shot 2 the man wears the glasses, in Shot 4 he does not. & Call 1: merged into one issue V2+V3 (the same discrepancy): ``The same DS man wears large-framed glasses in Shot 2 (frames 4, 5) and none in Shot 4 (frames 3, 4); no action of taking them off is visible between the shots.'' Call 2: confirmed, inconsistency confirmed, high confidence: his face is clearly visible in all compared frames, and no action or other normal cause explains the difference. Severity: moderate (one appearance detail; the man stays recognizable as the same person). $D = 25 \times 1 = 25$. \\
\midrule
V4 (unspecified-object check; inter-shot; object\_identity) & The trousers below the man's waist differ in the same washroom area marked by the white-framed mirror and the sinks: bright red in Shot 2, black in Shot 3. & Call 1: kept, clarified (the two visible pairs of trousers belong to different men). Call 2: rejected, not expected to match, high confidence: the red trousers are the DS man's and the black ones the other man's, so the difference is no evidence that one person's clothing changed. No deduction. \\
\bottomrule
\end{tabularx}
\end{table*}

\subsection{Confirmed issues by subcategory}
\label{sec:failure-types}

Table~\ref{tab:subcat-counts} counts the confirmed issues of each model by subcategory, and it is the basis of the failure types in Sec.~4.2.
In video-prompt alignment, missing required content (presence\_count) is the largest subcategory overall and dominates CineTrans-DiT.
Missed required cuts (shot\_structure) occur almost only for LongLive and CineTrans-DiT.
In intra-shot consistency, rendering instability is concentrated in CineTrans-DiT, and identity and appearance drift are concentrated in LongLive.
Across shots, changes of character appearance, which include worn items, are the most frequent subcategory for every model, and the sweep's \code{other} was never confirmed.
Over all eight models, 1,514 confirmed intra-video issues have an unspecified subject.
Of these, 1,248 are inter-shot, including 572 of character appearance (worn items), 329 of object identity, and 295 of setting and background.
The other 266 are intra-shot, including 133 of presence instability.

\begin{table*}[t]
\caption{\textbf{Confirmed issues by subcategory and model} (200 episodes per model).}
\label{tab:subcat-counts}
\centering
\scriptsize
\setlength{\tabcolsep}{3pt}
\begin{tabular*}{\textwidth}{@{\extracolsep{\fill}}lrrrrrrrrr@{}}
\toprule
Subcategory & Wan2.1 & Wan2.2 & LongLive & CineTrans-DiT & MultiShotMaster & HoloCine & ShotStream & StoryMem & All \\
\midrule
\multicolumn{10}{@{}l}{\textit{Video-prompt alignment}} \\
presence\_count & 163 & 56 & 944 & 3,132 & 750 & 690 & 612 & 103 & 6,450 \\
attribute & 463 & 257 & 408 & 652 & 362 & 195 & 379 & 149 & 2,865 \\
action\_state & 367 & 97 & 773 & 1,412 & 536 & 364 & 1,057 & 127 & 4,733 \\
interaction & 116 & 23 & 220 & 373 & 153 & 62 & 238 & 37 & 1,222 \\
spatial\_orientation & 339 & 134 & 697 & 1,053 & 359 & 329 & 480 & 139 & 3,530 \\
setting\_environment & 66 & 55 & 124 & 215 & 49 & 35 & 111 & 30 & 685 \\
camera\_composition & 159 & 69 & 618 & 802 & 150 & 165 & 251 & 84 & 2,298 \\
style\_rendering & 38 & 17 & 19 & 22 & 33 & 7 & 30 & 5 & 171 \\
stated\_continuity & 4 & 2 & 1 & 0 & 0 & 0 & 2 & 0 & 9 \\
shot\_structure & 0 & 0 & 121 & 76 & 0 & 1 & 0 & 0 & 198 \\
\midrule
\multicolumn{10}{@{}l}{\textit{Intra-shot consistency}} \\
identity\_drift & 6 & 2 & 56 & 4 & 1 & 4 & 0 & 1 & 74 \\
appearance\_drift & 28 & 13 & 125 & 15 & 10 & 6 & 2 & 15 & 214 \\
action\_discontinuity & 6 & 5 & 13 & 2 & 5 & 6 & 0 & 4 & 41 \\
object\_instability & 32 & 23 & 58 & 17 & 20 & 17 & 21 & 20 & 208 \\
presence\_instability & 33 & 12 & 106 & 18 & 24 & 16 & 20 & 21 & 250 \\
background\_drift & 8 & 4 & 21 & 1 & 4 & 1 & 2 & 5 & 46 \\
physical\_integrity & 2 & 0 & 28 & 15 & 7 & 7 & 1 & 6 & 66 \\
relation\_break & 2 & 0 & 4 & 0 & 0 & 1 & 2 & 0 & 9 \\
unintended\_cut & 17 & 18 & 46 & 5 & 24 & 36 & 3 & 27 & 176 \\
rendering\_instability & 2 & 2 & 4 & 114 & 6 & 20 & 8 & 10 & 166 \\
\midrule
\multicolumn{10}{@{}l}{\textit{Inter-shot consistency}} \\
character\_identity & 180 & 117 & 40 & 87 & 52 & 9 & 33 & 28 & 546 \\
character\_appearance & 435 & 358 & 189 & 278 & 131 & 71 & 140 & 107 & 1,709 \\
object\_identity & 182 & 175 & 63 & 64 & 64 & 21 & 67 & 69 & 705 \\
setting\_background & 156 & 180 & 46 & 13 & 42 & 17 & 34 & 52 & 540 \\
environment\_lighting & 54 & 62 & 1 & 3 & 19 & 6 & 8 & 23 & 176 \\
rendering\_style & 24 & 21 & 2 & 4 & 9 & 3 & 2 & 7 & 72 \\
instance\_count & 23 & 9 & 44 & 20 & 14 & 7 & 25 & 9 & 151 \\
state\_action\_continuity & 13 & 11 & 6 & 12 & 11 & 9 & 10 & 10 & 82 \\
spatial\_continuity & 3 & 3 & 1 & 2 & 3 & 2 & 3 & 2 & 19 \\
\bottomrule
\end{tabular*}
\end{table*}

\section{Limitations and Ethics}
\label{sec:limitations}

\paragraph{Detector recall and spatial judgment.}
Entity discovery is influenced by the recall of the detector (main paper, Sec.~5).
The detector may miss small, thin, or flat objects.
The background discovery step and the unspecified-object check (Sec.~\ref{sec:uoc}) find such objects only when their inventories list them, so inconsistencies of unlisted objects can go undetected.
Without 3D information, the agents also judge 3D spatial consistency less reliably (main paper, Sec.~5).

\paragraph{Verification by the same model.}
The verifier is independent of the context and reasoning of the detection call, but it uses the same model (Sec.~\ref{supp:sec:verification}).
The prompts of the episodes were written by GPT-6 Astra, a model of the same family as GPT-6 Luna.
Errors that these models make consistently are therefore not removed.
Replacing GPT-6 Luna with Gemini~3.8 Flash in the evaluation stages gives similar agreement with the raters and the same order of the five models (Sec.~\ref{sec:repeats}).

\paragraph{Identity of unprompted objects.}
Prompt binding and cross-shot matching decide whether an object is a contract entity or an unspecified one, and whether occurrences in different shots are the same entity.
These decisions determine whether an issue counts as specified or unspecified.

\paragraph{Temporal sampling.}
Six keyframes per shot, about one second apart, can miss brief artifacts.
They can also make a fast entry into view or an abrupt screen change look like an object popping in or an unexplained jump between keyframes.
The intra-shot dimension is affected most.

\paragraph{Scoring.}
The severity penalties, confidence factors, and equal weights are design choices, not fitted to human judgments.
Sec.~\ref{sec:sensitivity} tests alternative values, and Sec.~\ref{sec:length} discusses dividing the deduction by the number of shots.

\paragraph{Contracts and scope.}
The contracts were compiled and reviewed by language models under written rules and program checks, with sample checks by the authors (Sec.~\ref{sec:contract-format}).
The 200 episodes come from two source indices of narrative film.
The human study covers 24 episodes and five of the eight models, and it measures agreement on whole videos, not on entities.

\paragraph{Ethics.}
The benchmark consists of text: prompts written by a language model from frames of the source videos, and contracts compiled from these prompts.
Evaluation uses only videos generated from the prompts.
Prompts describe the visible appearance of people, including apparent age, gender, and skin tone, as written from the source frames.
The evaluator uses these descriptions only to check whether a generated character matches its description and stays the same across shots.
The human study asked raters to judge generated videos and their prompts, and the analysis uses only the ratings, the time spent per episode, and pseudonymous rater labels.
Raters took part voluntarily and were told when recruited that their ratings would be used, without names, to evaluate the benchmark in a research paper.

\paragraph{Release.}
The following will be released:
(i) the prompts of the 200 episodes, their contracts in both versions (with and without the supporting prompt quotes), the compiled requirement lists, and the written compilation rules with their program checks;
(ii) the evaluator code with the verbatim agent prompts and output schemas, of which the tables of Sec.~\ref{sec:evaluator} give condensed forms;
(iii) the configuration of every reported run: model identifiers, reasoning effort, and run dates;
(iv) the stored outputs of every evaluated video in the reported runs: entity ledgers, detection candidates, verification verdicts, confirmed issues with their severities, and scores;
(v) the per-video mean ratings of the human study for the three questions and the overall rating;
(vi) the generated videos of the eight models, subject to each video model's license.
The source videos are \mbox{not part of the release.}

\end{document}